\pdfoutput=1
\documentclass[letterpaper, 10 pt, conference]{ieeeconf}  

\IEEEoverridecommandlockouts

\usepackage{moreverb,url}

\usepackage[colorlinks,bookmarksopen,bookmarksnumbered,citecolor=red,urlcolor=red]{hyperref}
\usepackage{epstopdf}

\usepackage[activate=alltext]{microtype}
\usepackage{ifpdf}
\usepackage{times}
\usepackage{graphicx}
\usepackage{color}
\usepackage{amsmath}
\usepackage{amssymb}
\usepackage{algorithmic}
\usepackage{algorithm}
\usepackage{hhline}
\usepackage{rotating}
\usepackage{multirow}
\usepackage{url}
\usepackage{comment}
\usepackage{array}

\let\oldemptyset\emptyset
\let\emptyset\varnothing

\makeatletter
\let\NAT@parse\undefined
\makeatother
\usepackage[numbers, sort]{natbib}

\newcommand{\ignore}[1]{}

\newcommand{\secref}[1]{Section~\ref{#1}}

\renewcommand{\eqref}[1]{Equation~(\ref{#1})}
\newcommand{\figref}[1]{Figure~\ref{#1}}
\newcommand{\tabref}[1]{Table~\ref{#1}}

\newcounter{extension}
\DeclareMathOperator*{\argmin}{argmin}

\newcommand{\objects}{\mathcal{O}}
\newcommand{\containers}{\mathcal{C}}
\newcommand{\pairs}{\mathcal{P}}
\newcommand{\ratingMat}{\mathbf{R}}
\newcommand{\biasMat}{\mathbf{B}}
\newcommand{\users}{\mathcal{U}}
\newcommand{\biasPair}{b_i}
\newcommand{\biasUser}{b_j}

\newcommand{\videourl}{\url{http://www.informatik.uni-freiburg.de/\%7Eabdon/task_preferences.html}}

\hypersetup{
  colorlinks,
    citecolor=black,
    filecolor=black,
    linkcolor=black,
    urlcolor=black
}

\begin{document}

\title{\LARGE \bf   {Collaborative Filtering for Predicting\\User Preferences 
  for Organizing Objects} }
\author{Nichola Abdo* \and Cyrill Stachniss\textsuperscript{\ddag} \and Luciano 
  Spinello* \and Wolfram  Burgard* \thanks{\hspace{-\parindent}* The University 
    of Freiburg, Department of Computer Science,  79110 Freiburg, Germany.}  
  \thanks{\textsuperscript{\hspace{-\parindent}\ddag{}} The University of Bonn, 
    Institute of Geodesy and Geoinformation, 53115 Bonn, Germany} }

\maketitle

\begin{abstract}
  As service robots become more and more capable of performing useful
  tasks for us, there is a growing need to teach robots how we expect
  them to carry out these tasks. However, different users typically
  have their own preferences, for example with respect to arranging
  objects on different shelves.  As many of these preferences depend
  on a variety of factors including personal taste, cultural
  background, or common sense, it is challenging for an expert to pre-program a 
  robot in order to accommodate all potential users. At the
  same time, it is impractical for robots to constantly query users
  about how they should perform individual tasks. In this work, we
  present an approach to learn patterns in user preferences for the
  task of tidying up objects in containers, e.g., shelves or
  boxes. Our method builds upon the paradigm of collaborative
  filtering for making personalized recommendations and relies on data
  from different users that we gather using crowdsourcing.  To deal
  with novel objects for which we have no data, we propose a method
  that compliments standard collaborative filtering by leveraging
  information mined from the Web. When solving a tidy-up task, we
  first predict pairwise object preferences of the user.  Then, we
  subdivide the objects in containers by modeling a spectral
  clustering problem.  Our solution is easy to update, does not
  require complex modeling, and improves with the amount of user data.
  We evaluate our approach using crowdsourcing data from over 1,200
  users and demonstrate its effectiveness for two tidy-up scenarios.
  Additionally, we show that a real robot can reliably predict user
  preferences using our approach.
  \end{abstract}

\section{Introduction}
\label{sec:intro}
\begin{figure*}
\setlength{\fboxsep}{0pt}
 \centering
~\hfill \fbox{\includegraphics[height=4cm]{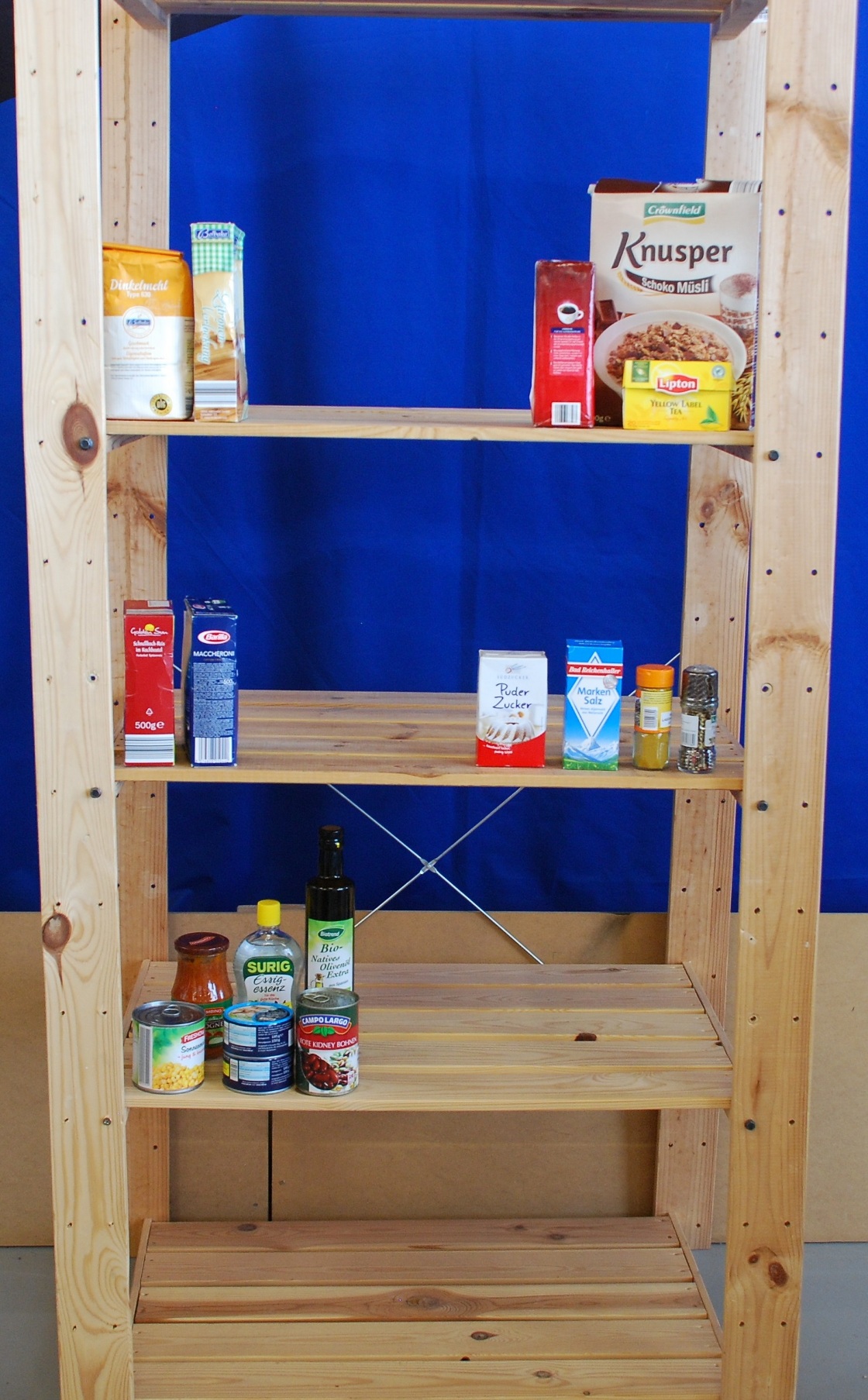}}
\fbox{\includegraphics[height=4cm]{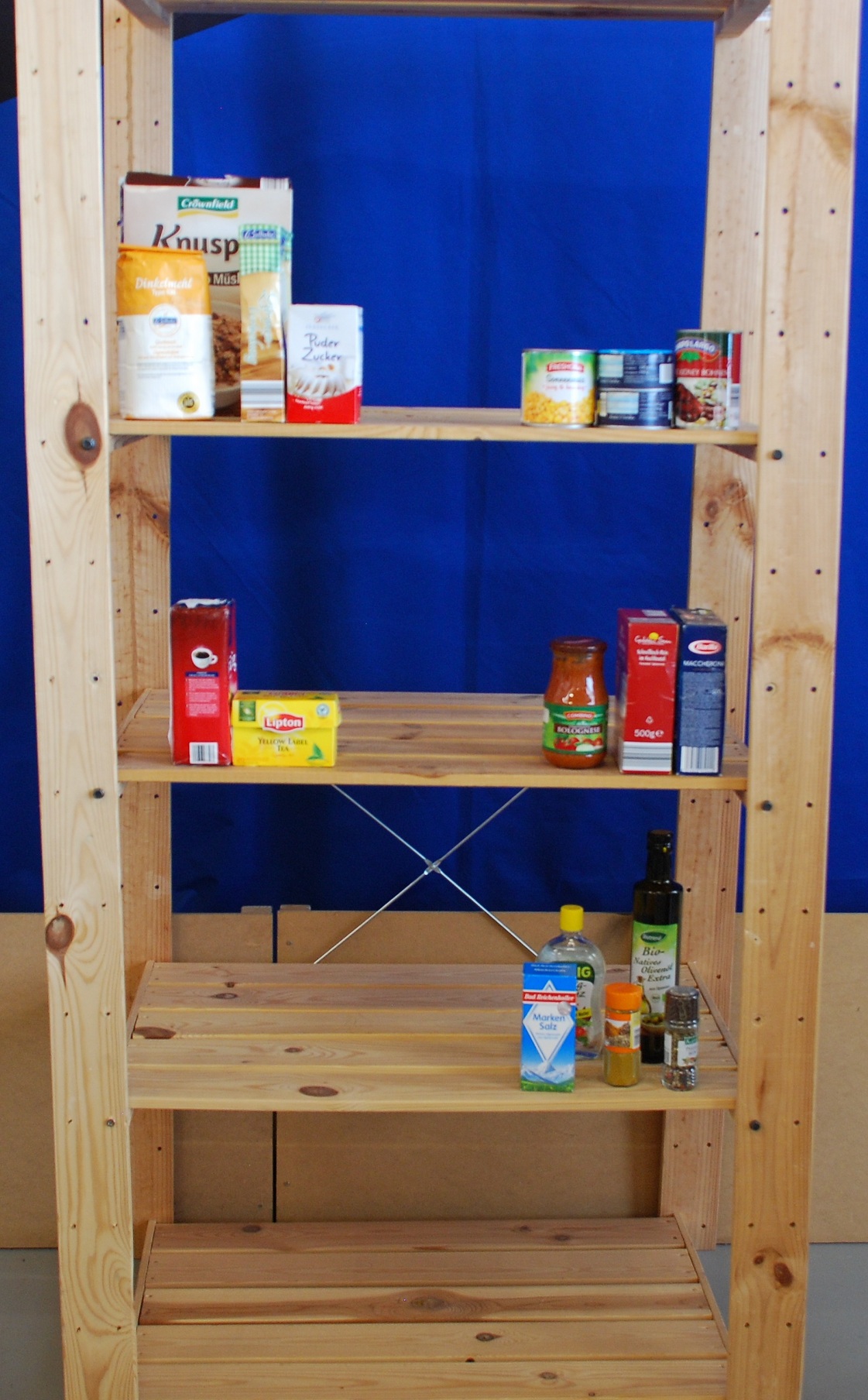}}
\fbox{\includegraphics[height=4cm]{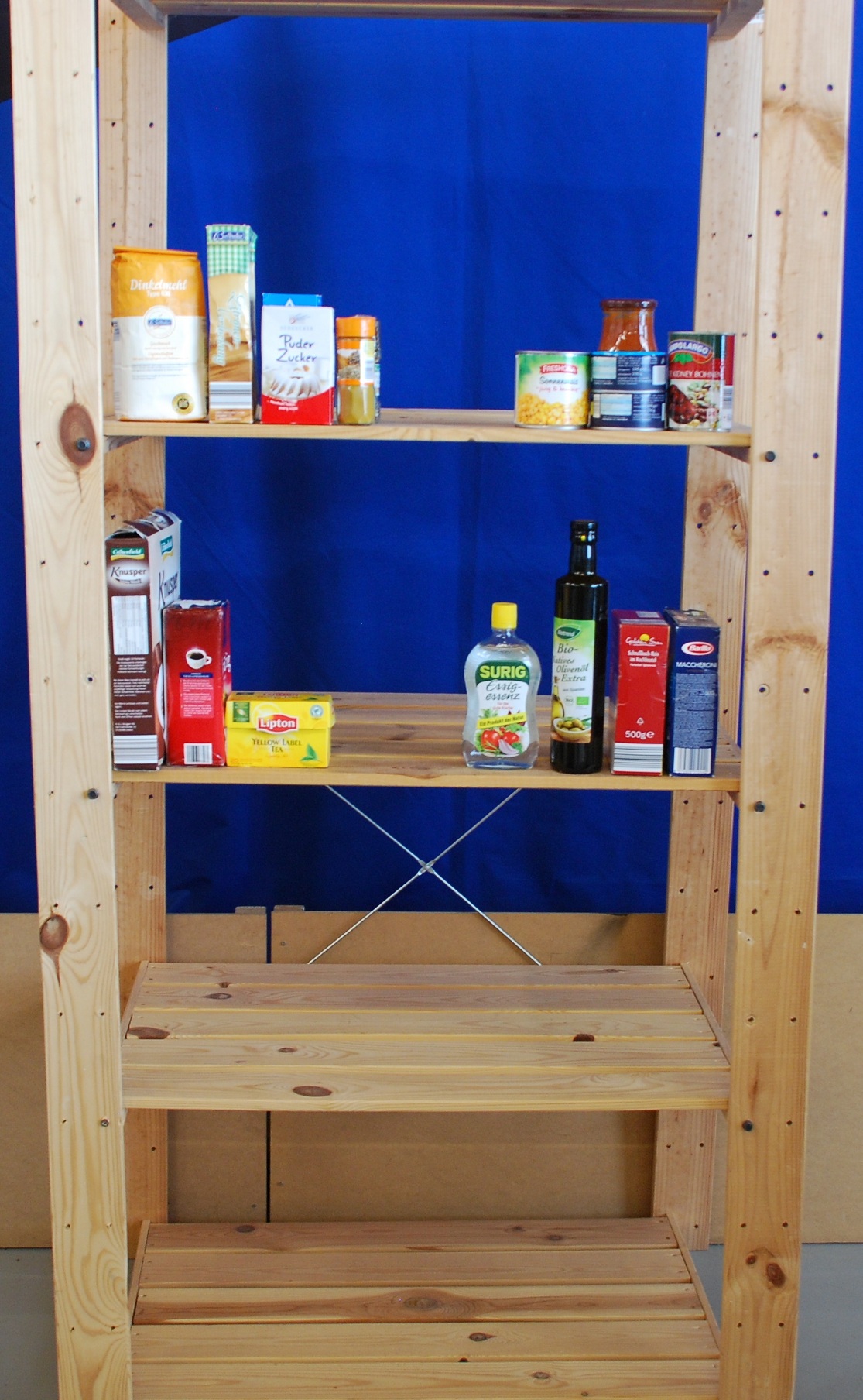}}\hfill
\fbox{\includegraphics[height=4cm]{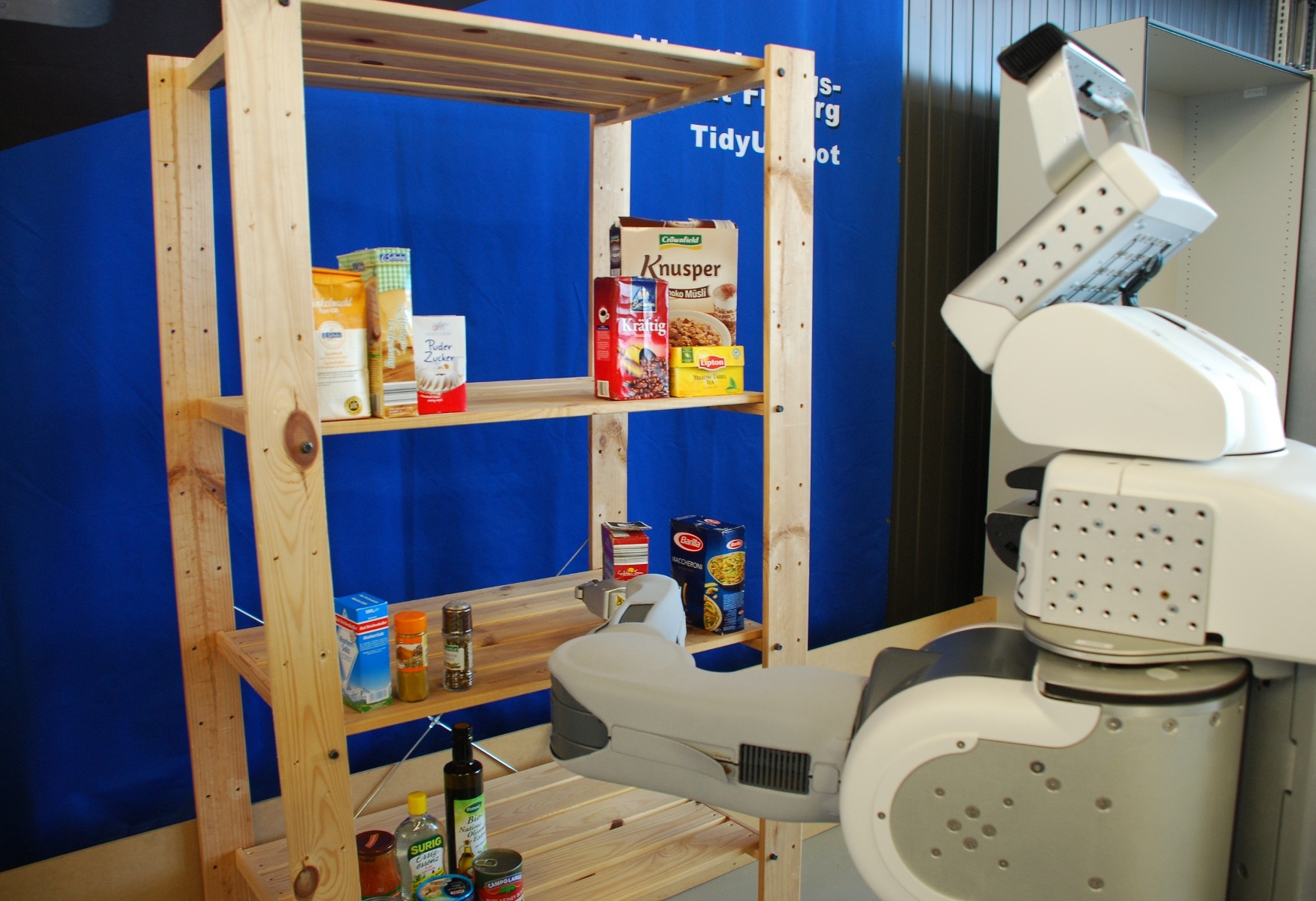}}\hfill~
\caption{Left: different ways of organizing a set of grocery objects on shelves 
  according to varying user preferences. Right: our approach enables a service 
    robot to tidy up objects by predicting and following such subjective 
    preferences.  We predict pairwise preferences between objects with respect 
    to placing them on the same shelf.  We then assign objects to different 
    shelves by maximally satisfying these preferences.}
\label{fig:motivation}
\end{figure*}

One of the key goals of robotics is to develop autonomous service
robots that assist humans in their everyday life. One envisions smart
robots that can undertake a variety of tasks including tidying up,
cleaning, and attending to the needs of disabled people. For
performing such tasks effectively, each user should teach her robot
\emph{how} she likes those tasks to be performed. However, learning
user preferences is an intricate problem. In a home scenario, for
example, each user has a preferred way of sorting and storing
groceries and kitchenware items in different shelves or
containers. Many of our preferences stem from factors such as personal
taste, cultural background, or common sense, which are hard to
formulate or model a priori. At the same time, it is highly
impractical for the robot to constantly query users about their
preferences.

In this work, we provide a novel solution to the problem of learning
user preferences for arranging objects in tidy-up tasks. Our method is
based on the framework of collaborative filtering, which is a popular
paradigm from the data-mining community. Collaborative filtering is
generally used for learning user preferences in a wide variety of
practical applications including suggesting movies on Netflix or
products on Amazon. Our method predicts user preferences of pairwise
object arrangements based on partially-known preferences, and then computes the 
best subdivision of objects in shelves or boxes. It is able
to encode multiple user preferences for each object and it does not
require that all user preferences are specified for all object-pairs.
Our approach is even able to make predictions when novel objects,
unknown to previous users, are presented to the robot. For this, we
combine collaborative filtering with a mixture of experts that compute
similarities between objects by using object hierarchies.  These
hierarchies consist of product categories downloaded from online
shops, supermarkets, etc. Finally, we organize objects in different
containers by finding object groups that maximally satisfy the
predicted pairwise constraints. For this, we solve a minimum $k$-cut
problem by efficiently applying self-tuning spectral clustering. Our
prediction model is easy to update and simultaneously offers the
possibility for lifelong learning and improvement.

To discover patterns in user preferences, we first bootstrap our learning by 
collecting many user preferences, e.g., through crowdsourcing surveys. Using 
this data, we learn a model for object-pair preferences for a certain tidy-up
task. Given partial knowledge of a new user's preferences (e.g., by querying the 
    user or observing how the user has arranged some objects in the 
    environment), the robot can then use this model to predict unknown 
object-pair preferences of the new user, and sort objects accordingly.

To summarize, we make the following contributions:
\begin{itemize}
\item We model the problem of organizing objects in different
  containers using the framework of collaborative filtering for
  predicting personalized preferences;
\item We present an approach by which a service robot can easily learn
  the preferences of a new user using observations from the
  environment and a model of preferences learned from several previous
  users;
\item We present a novel method to complement standard collaborative
  filtering techniques by leveraging information from the Web in cases
  where there is not enough ratings to learn a model;
\item We present an extensive experimental evaluation using
  crowdsourcing data that demonstrates that our approach is suitable
  for lifelong learning of user preferences with respect to organizing
  objects.
\end{itemize}

Our evaluation covers two relevant tidy-up scenarios, arranging toys
in different boxes and grocery items on shelves, as well as a
real-robot experiment.  For training, we collected preferences from
over 1,200 users through different surveys.

This paper incorporates the approach and initial results from our
previous conference publication~\cite{abdo15icra}, and extends our
work in the following ways: \emph{i}) we present a more thorough
review of related work, \emph{ii}) we present a new extension of our
approach for inferring the preferences of new users in an efficient
manner, and \emph{iii}) we conduct a more extensive experimental
evaluation of all aspects of our method, presenting new results and
insights.

\section{Related Work}
\label{sec:related}
Equipping service robots with the knowledge and skills needed to attend to 
complex chores in domestic environments has been the aim of researchers for 
years. Indeed, recent advances in perception, manipulation, planning, and 
control have enabled robots to perform a variety of chores that range from 
cleaning and tidying up to folding 
laundry~\cite{saxena2008robotic,hess12iros,miller2012geometric,doumanoglou2014autonomous}.  
However, as highlighted by a number of researchers, service robots should also 
be able to attend to such tasks in a manner that corresponds to the personal 
preferences of end users~\cite{TakayamaChores13,Forlizzi2006, 
  Dautenhahn2005IROS,PantofaruHRI12,Ray2008, Smarr14IJSR, Cha2015HRI}.  For 
  example, the results of \citeauthor{PantofaruHRI12} show that people exhibit 
  strong feelings with respect to robots organizing personal items, suggesting 
  the need for the robot to ask humans to make decisions about where to store 
  them~\cite{PantofaruHRI12}. In this work, we present a novel approach by which 
  robots can discover patterns in organizing objects from a corpus of user 
  preferences in order to achieve preferred object arrangements when tidying up 
  for a specific user. This allows a robot to predict the preferred location 
  (e.g., a specific shelf) to store an object by observing how the user has
  previously arranged other objects in the same environment. Several researchers 
  have leveraged the fact that our environments are rich with cues that can 
  assist robots in various tasks that require reasoning about objects and their 
  locations. For example, different works have addressed object classification 
  or predicting the locations of objects using typical 3D structures in indoor 
  environments or object-object relations such as co-occurrences in a 
  scene~\cite{joho11ras, aydemir2012exploiting,lorbach_object_search_2014,
    icra14ensmln, Kunze14}. However, our work is concerned with learning 
    pairwise object preferences to compute preferred arrangements when tidying 
    up. In the remainder of this section, we discuss prior work in the 
    literature that is most relevant to the problem we address and the 
    techniques we present.
\paragraph{Learning Object Arrangements and Placements} Recently, 
  \citeauthor{schuster2010perceiving} presented an approach for distinguishing 
  clutter from clutter-free areas in domestic environments so that a robot can 
  reason about suitable surfaces for placing 
  objects~\cite{schuster2010perceiving}. Related to that, the work of   
  \citeauthor{jiang2012learning} targets learning physically stable and 
  semantically preferred poses for placing objects given the 3D geometry of the 
  scene~\cite{jiang2012learning}. \citeauthor{joho12rss} developed a novel 
  hierarchical nonparametric Bayesian model for learning scenes consisting of 
  different object constellations~\cite{joho12rss}. Their method can be used to 
  sample missing objects and their poses to complete partial scenes based on 
  previously seen constellations. Other approaches have targeted synthesising 
  artificial 3D object arrangements that respect constraints like physical 
  stability or that are semantically plausible~\cite{xu2002constraint, 
    Fisher:2012:ESO:2366145.2366154}. We view such works as complimentary to 
    ours, as we address the problem of learning preferred groupings of objects 
    in different containers (e.g., shelves) for the purpose of tidying up.  
    After predicting the preferred container for a specific object, our approach 
    assumes that the robot is equipped with a suitable technique to compute a 
    valid placement or pose of the object in that location.  Moreover, as we 
    explicitly consider sorting objects when tidying up, we do not reason about 
    object affordances associated with various human poses and activities in the 
    scene (e.g., cooking, working at a desk, etc) when computing arrangements, 
    as in the work of \citeauthor{jiang2012humancontext} and    
    \citeauthor{savva2014scenegrok}~\cite{jiang2012humancontext, 
      savva2014scenegrok}.
    
      Related to our work, previous approaches have addressed learning 
      organizational patterns from surveys conducted with different users.  
      \citeauthor{Schuster12} presented an approach for predicting the location 
for storing different objects (e.g., cupboard, drawer, fridge, etc) based on 
other objects observed in the environment~\cite{Schuster12}.  They consider 
different features that capture object-related properties (e.g., the purpose of 
    an object or its semantic similarity to other objects) and train classifiers 
that predict the location at which an object should be stored.  Similarly, 
     \citeauthor{Cha2015HRI} explored using different features describing both 
     objects and users to train classifiers for predicting object locations in 
     user homes~\cite{Cha2015HRI}. Similar to \citeauthor{Schuster12}, we
    also make use of a similarity measure based on hierarchies
   mined from the Web and use it for making predictions for
  objects for which we have no training data. However, in contrast to these 
  works, our approach learns latent organizational patterns across different 
  users in a collaborative manner and without the need for designing features 
  that describe objects or users.  Recently, \citeauthor{toris2015unsupervised} 
  presented an approach to learn placing locations of objects based on 
  crowdsourcing data from many users~\cite{toris2015unsupervised}. Their 
  approach allows for learning multiple hypotheses for placing the same object, 
  and for reasoning about the most likely frame of reference when learning the 
  target poses.  They consider different pick-and-place tasks such as setting a 
  table or putting away dirty dishes, where the aim is to infer the final object 
  configuration at the goal.  Our approach is also able to capture multiple 
  modes with respect to the preferred location for placing a certain object.  In 
  contrast to \citeauthor{toris2015unsupervised}, we explicitly target learning 
  patterns in user preferences with respect to sorting objects in different 
  containers.  Moreover, in contrast to the above works, our method allows the 
  robot to adapt to the available number of containers in a certain environment 
  to sort the objects while satisfying the user's preferences as much as 
  possible.  Note that in this work, we assume the robot is equipped with a map 
  of the environment where relevant containers are already identified.  Previous 
  work has focused on constructing such semantic maps that are useful for robots 
  when planning to solve complex household 
  tasks~\cite{Vasudevan2007,zender2008conceptual,iros12semantic_mapping}.
  
      \paragraph{Service Robots Leveraging the Web} Recently, several researches 
      have leveraged the Web as a useful source of information for assisting 
      service robots in different tasks~\cite{tenorth11www, kehoe2013ICRA}.
To cope with objects that are not in the robot's database, our method combines 
collaborative filtering with a mixture of experts approach based on object 
hierarchies we mine from online stores. This allows us to compute the semantic 
similarity of a new object to previously known objects to compensate for missing 
user ratings. The work by~\citeauthor{Schuster12} has also utilized such 
similarity measures as features when training classifiers for predicting 
locations for storing objects~\cite{Schuster12}. \citeauthor{irosws11germandeli} 
also leverage information from online stores but in the context of object 
detection~\cite{irosws11germandeli}. \citeauthor{Kaiser2014} recently presented 
an approach for mining texts obtained from the Web to extract common sense 
knowledge and object locations for planning tasks in domestic 
environments~\cite{Kaiser2014}. Moreover, \citeauthor{icra14ensmln} presented an 
ensemble approach where different perception techniques are combined in the 
context of detecting everyday objects~\cite{icra14ensmln}.

\paragraph{Collaborative Filtering}
We predict user preferences for organizing objects based on the
framework of collaborative filtering, a successful paradigm from the data mining 
community for making personalized user
recommendations of 
products~\cite{CannyCF2002,bennett2007netflix,koren2008factorization,koren2010factor,sarwar2001item}.  
Such techniques are known for their scalability and suitability for life-long 
learning settings, where the quality of the predictions made by the 
recommender system improves with more users providing their ratings. Outside 
the realm of customers and products, factorization-based collaborative filtering 
has recently been successfully applied to other domains including 
action-recognition in videos \cite{Matikainen2012ModelRecom} and 
predicting drug-target interactions~\cite{Temerinac-Ott2015}. Recently, 
           \citeauthor{Matikainen2013Bandits} combined a recommender system 
           with a multi-armed bandit formulation for suggesting good floor 
           coverage
strategies to a vacuum-cleaning robot by modeling different room layouts 
as users~\cite{Matikainen2013Bandits}. To the best of our knowledge, we 
believe we are the first work to use collaborative filtering for 
predicting personalized user preferences in the context of service robotics.

\paragraph{Crowdsourcing for Robotics}
To learn different user preferences, we collect data from many non-expert users 
using a crowdsourcing platform. Prior work has also leveraged crowdsourcing for 
data labeling or as an efficient platform for transferring human knowledge to 
robots, e.g.,~\cite{DengKrauseFei-Fei_CVPR2013,kent2015icra}. For example, 
  \citeauthor{Sorokin2010} utilized crowdsourcing to teach robots how to grasp 
  new objects~\cite{Sorokin2010}.  Moreover, several researchers have used 
  crowdsourcing to facilitate learning manipulation tasks from large numbers of 
  human 
  demonstrations~\cite{chungaccelerating,toris2014robot,toris2015unsupervised,ratner2015web}.  
  In the context of learning user preferences, \citeauthor{jain2015icra} 
  recently presented a new crowdsourcing platform where non-experts can label 
  segments in robot trajectories as desirable or not~\cite{jain2015icra}. This 
  is then used to learn a cost function for planning preferred robot 
  trajectories in different indoor environments.

\section{Collaborative Filtering for Predicting \\
  Pairwise Object Preferences}
\label{sec:CF}
Our goal is to enable a service robot to reason about the preferred way to sort 
a set of objects into containers when tidying up in the environment of a 
specific user. To achieve this, we aim at predicting the preferences of the user 
with respect to grouping different objects together. As the types of objects 
(e.g., grocery items) and number of containers (e.g., shelves) typically vary 
across environments, we aim to learn user preferences for object-object 
combinations, rather than directly learning an association between an object and 
a specific container.  The problem of predicting an object-object preference for 
a user closely resembles that of suggesting products to customers based on their 
tastes. This problem is widely addressed by employing recommender systems, 
  commonly used by websites and online stores such as Amazon and Netflix. The 
  key idea there is to learn to make recommendations based on the purchasing 
  histories of different users collaboratively.
In the same spirit of reasoning about products and users, our method relates 
pairs of objects to users. We predict a user preference, or \textit{rating}, for 
an object-pair based on two sources of information: \emph{i}) known preferences 
of the user, e.g., how the user has previously organized other objects, and 
\emph{ii}) how other users have organized these objects in their environments.

\subsection{Problem Formulation}
\label{sec:probForm}

\begin{figure}[t]
 \centering
\includegraphics[height=4.5cm]{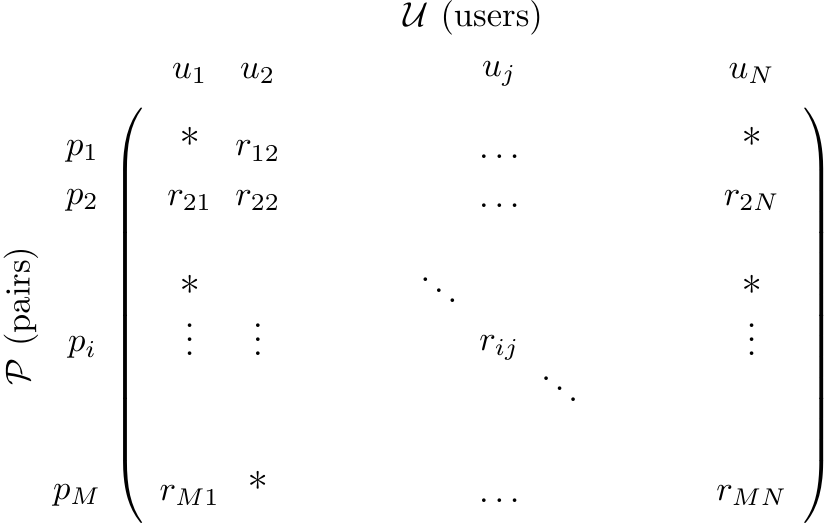}
\caption{The ratings matrix $\ratingMat$. Each entry $r_{ij}$
  corresponds to the rating of a user $u_j$ for an object-pair
  $p_i=\{o_k, o_l\}$, a value between 0 and 1 denoting whether the two objects 
  should be placed in the same container or not. Our goal is to predict the
  missing ratings denoted by * and compute a partitioning of the objects in 
  different containers that satisfies the user preferences.}
\label{fig:ratingsMatrix}
\end{figure}

More formally, let $\objects = \{o_1, o_2, \dots, o_O\}$ be a set of objects, 
     each belonging to a known class, e.g., book, coffee, stapler, etc.  
     Accordingly, we define $\pairs=\{p_1, p_2, \dots, p_M\}$ as the set of all 
     pairs of objects from $\objects$. We assume to have a finite number of 
     \emph{containers} $\containers=\{c_1, c_2, \dots, c_C\}$, which the robot 
     can use to organize the objects, e.g., shelves, drawers, boxes, etc. We 
     model each container as a set which could be $\oldemptyset$ or could 
     contain a subset of $\objects$.  Given a set of users $\users = \{u_1, 
     \dots, u_N\}$, we assign a rating $r_{ij}$ to a pair $p_i = \{o_l, o_k\}$ 
     to denote the preference of user $u_j$ for placing $o_l$ and $o_k$ in the 
     same container. Each rating takes a value between 0 and 1, where 0 means 
     that the user prefers to place the objects of the corresponding pair into 
     separate containers, and 1 means that the user prefers placing them 
     together. For convenience, we use $r(o_l, o_k)$ to denote the rating for 
     the pair consisting of objects $o_l$ and $o_k$ when the user is clear from 
     the context. We can now construct a ratings matrix $\ratingMat$ of size $M 
     \times N$, where the rows correspond to the elements in $\pairs$ and the 
     columns to the users, see \figref{fig:ratingsMatrix}.  We use $R$ to denote 
     the number of known ratings in $\ratingMat$. Note that typically, $R \ll 
     MN$, i.e., $\ratingMat$ is missing most of its entries.  This is due to the 
     fact that each user typically ``rates'' only a small subset of 
     object-pairs. In this work, we denote the set of indices of object-pairs 
     that have been rated by user $u_j$ by $\mathcal{I}_j \subseteq \{1, \dots, 
     M\}$.  Analogously, $\mathcal{J}_i \subseteq \{1, \dots, N\}$ is the set of 
     indices of users who have rated object-pair $p_i$.
     
     Given a set of objects $\objects'\subseteq\objects$ that the robot has to 
     sort for a specific user $u_j$, and the set of containers $\containers$ 
     available for the robot to complete this task, our goal is to: \emph{i})
    predict the unknown preference $\hat{r}_{ij}$ of the user for each of the 
    object-pairs $\pairs'$ over $\objects'$ and, accordingly, \emph{ii}) assign 
    each object to a specific container such that the user's preferences are 
    maximally satisfied.
   
\subsection{Collaborative Learning of User Preferences}
\label{sec:cfLearning}

We aim to discover latent patterns in the ratings matrix $\ratingMat$ that 
enable us to make predictions about the preferences of users. For this, we take 
from factorization-based collaborative
filtering~\cite{koren2008factorization,koren2010factor}.
First, we decompose $\ratingMat$ into a bias matrix $\biasMat$ and a residual 
ratings matrix $\overline{\ratingMat}$:
\begin{equation}
\label{eq:rDecomp}
\ratingMat = \biasMat + \overline{\ratingMat}.
\end{equation} 
Each entry $b_{ij}$ in $\biasMat$ is formulated as follows:
\begin{equation}
b_{ij} = \mu + \biasPair + \biasUser,
\end{equation}
where $\mu$ is a global bias term, $\biasPair$ is the bias of the pair
$p_i$, and $\biasUser$ is the bias of user $u_j$. We compute $\mu$ as the
mean rating over all users and object-pairs in $\ratingMat$, i.e.,
\begin{equation}
\mu = \frac{1}{R} \sum_{i=1}^M \sum_{j\in \mathcal{J}_i} r_{ij}.
\end{equation}
The bias $\biasUser$ describes how high or low a certain user $u_j$ tends to
rate object-pairs compared to the average user.  Similarly,
$\biasPair$ captures the tendency of a pair $p_i$ to receive high or low
ratings. For example, the pair $\{$\emph{salt}, \emph{pepper}$\}$
tends to receive generally high ratings compared to the pair $\{$\emph{candy}, 
      \emph{vinegar}$\}$.

After removing the bias, the residual ratings matrix
$\overline{\ratingMat}$ captures the fine, subjective user preferences that we 
aim to learn by factorizing the matrix to uncover latent patterns. Due to the 
large amount of missing ratings in $\overline{\ratingMat}$, it is infeasible to 
apply classical factorization techniques such as singular value decomposition.  
Instead, we learn a data-driven factorization based only on the \emph{known 
  entries} in $\overline{\ratingMat}$.  This approach has been shown to lead to 
  better results in matrix completion or factorization problems compared to 
  imputation of the missing values~\cite{CannyCF2002, koren2008factorization}.  
  We express $\overline{\ratingMat}$ as the product of an object-pair factors 
  matrix $\mathbf{S}^T$, and a user factors matrix $\mathbf{T}$ of sizes $M 
  \times K$ and $K \times N$, respectively.  Each column $\mathbf{s}_i$ of 
  $\mathbf{S}$ is a $K$-dimensional factors vector corresponding to an 
  object-pair $p_i$.  Similarly, each column $\mathbf{t}_j$ in $\mathbf{T}$ is a 
  $K$-dimensional factors vector associated with a user $u_j$.  We compute the 
  residual rating $\overline{r}_{ij}$ as the dot product of the factor vectors 
  for object-pair $p_i$ and user $u_j$, i.e.,

\begin{equation}
\label{eq:ratingDecomp}
\overline{r}_{ij} \,=\, \mathbf{s}_i^T \cdot \mathbf{t}_j.
\end{equation}
The vectors $\mathbf{s}$ and $\mathbf{t}$ are low-dimensional projections of the 
pairs and users, respectively, capturing latent characteristics of both. Pairs 
or users that are close to each other in that space are similar with respect to 
some property.  For example, some users could prefer to group objects together 
based on their shape, whereas others do so based on their function.

Accordingly, our prediction $\hat{r}_{ij}$ for the rating of an object-pair 
$p_i$ by a user $u_j$ is expressed as
\begin{equation}
\label{eq:ratingDetailed}
\begin{aligned}
\hat{r}_{ij} &= b_{ij} + \overline{r}_{ij}\\
    &= \mu + \biasPair + \biasUser + \mathbf{s}_i^T \cdot \mathbf{t}_j.
\end{aligned}
\end{equation}
We learn the biases and factor vectors from all available ratings in
$\ratingMat$ by formulating an optimization problem. The goal is to
minimize the difference between the observed ratings $r_{ij}$ made by
users and the predictions $\hat{r}_{ij}$ of the system over all known
ratings. Let the error associated with rating $r_{ij}$ be
\begin{equation}
\label{eq:predError}
e_{ij} = r_{ij} - (\mu + \biasPair + \biasUser + \mathbf{s}_i^T \cdot \mathbf{t}_j).
\end{equation}
We jointly learn the biases and factors that minimize the error over all
known ratings, i.e.,
\begin{equation}
\begin{aligned}
 b_*, \mathbf{S}, \mathbf{T} &= \argmin_{b_*,\mathbf{S},\mathbf{T}} \sum_{i=1}^M 
 \sum_{j\in \mathcal{J}_i} (e_{ij})^2 +\\
     &\frac{\lambda}{2}(\biasPair^2 + \biasUser^2 + \|\mathbf{s}_i\|^2 + 
         \|\mathbf{t}_j\|^2),
\end{aligned}
\label{eq:optimization}
\end{equation}
where $b_*$ denotes all object-pair and user biases and $\lambda$ is a 
regularizer.
To do so, we use L-BFGS optimization with a random initialization for all 
variables~\cite{nocedal1980lbfgs}. At every step of the optimization, we update 
the value of each variable based on the error gradient with respect to that 
variable, which we derive from~\eqref{eq:optimization}. 

\subsection{Probing and Predicting for New Users}
\label{sec:cfNewUsers}

After learning the biases and factor vectors for all users and object-pairs as 
in \secref{sec:cfLearning}, we can use \eqref{eq:ratingDetailed} to predict the 
requested rating $\hat{r}_{ij}$ of a user $u_j$ for an object-pair $p_i$ that 
she has not rated before. However, this implies that we have already learned the 
bias $\biasUser$ and factor vector $\mathbf{t}_j$ associated with that user. In other 
words, at least one entry in the $j$-th column of $\ratingMat$ should be known.
The set of known preferences for a certain user, used for learning her model, 
    are sometimes referred to as \emph{probes} in the recommender system 
    literature. In this work, we use \emph{probing} to refer to the process of 
    eliciting knowledge about a new user.

\subsubsection{Probing}
\label{sec:probing}
\begin{figure}
 \centering
\includegraphics[]{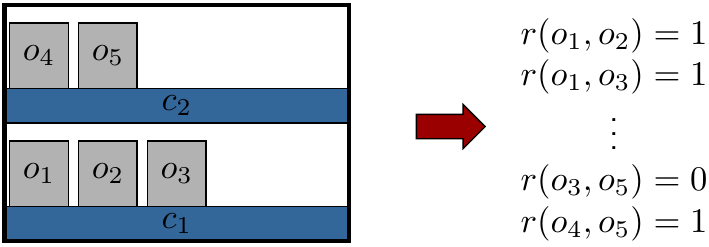}
\caption{A simple illustration of the probing process by which the robot can 
  infer some preferences for a new user. We set a rating of 0 for a pair of 
    objects that the robot observes to be in different containers, and a rating 
    of 1 for those in the same container. Using these ratings, we can learn a 
    model of the user to predict her preferences for other object-pairs.}
\label{fig:probingExample}
\end{figure}

In the context of a tidy-up service robot, we envision two strategies to do
so.  In the first probing approach, the robot infers some preferences of the 
user based on how she has previously sorted objects in the containers 
$\containers$ in the environment. By detecting the objects it encounters there, 
  the robot can infer the probe rating for a certain object-pair based on 
  whether the two objects are in the same container or not:
\begin{equation}
    r_{ij} = \begin{cases}
        1,& \text{if } o_l, o_k \in c_m\\
              0,& \text{if } o_l \in c_m, o_k\in c_n, m\neq n.
       \end{cases}
\label{eqn:probing}
\end{equation}
We do this for all object-pairs that the robot observes in the environment and 
fill the corresponding entries in the user's column with the inferred ratings, 
     see~\figref{fig:probingExample}.

In the second probing approach, we rely on actively querying the user
about her preferences for a set of object-pairs. For this, we rely on simple, 
      out-of-the-box user interface solutions such as a text interface where the 
      user can provide a rating. Let $P$ be the
maximum number of probe ratings for which the robot queries the user.
One naive approach is to acquire probes by randomly querying the user about $P$ 
object-pairs. However, we aim at making accurate predictions with as few probes 
as possible. Thus, we propose an efficient strategy based on insights into the 
factorization of \secref{sec:cfLearning}. The columns of the matrix $\mathbf{S}$ 
can be seen as a low dimensional projection of the rating matrix capturing the 
similarities between object-pairs; object-pairs that are close in that space 
tend to be treated similarly by users. We therefore propose to cluster the 
columns of $\mathbf{S}$ in $P$ groups, randomly select one column as a 
representative from each cluster, and query the user about the associated 
object-pair. For clustering, we use $k$-means with $P$ clusters. In this way, 
  the queries to the users are selected to capture the complete spectrum of 
  preferences.

  Note that the nature of a collaborative filtering system allows us to 
  continuously add probe ratings for a user in the ratings matrix, either 
  through observations of how objects are organized in the environment, or by 
  active querying as needed. This results in a life-long and flexible approach 
  where the robot can continuously update its knowledge about the user.

\begin{figure*}[t]
\centering
 ~\hfill\includegraphics[width=0.4\textwidth]{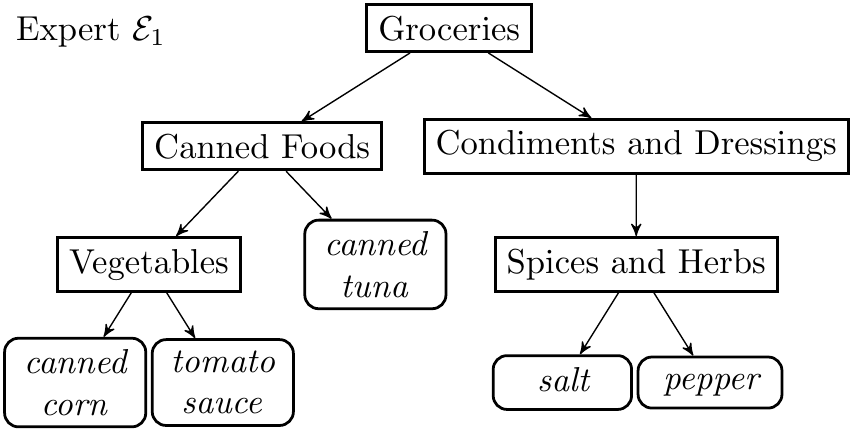}\hfill
  \includegraphics[width=0.38706\textwidth]{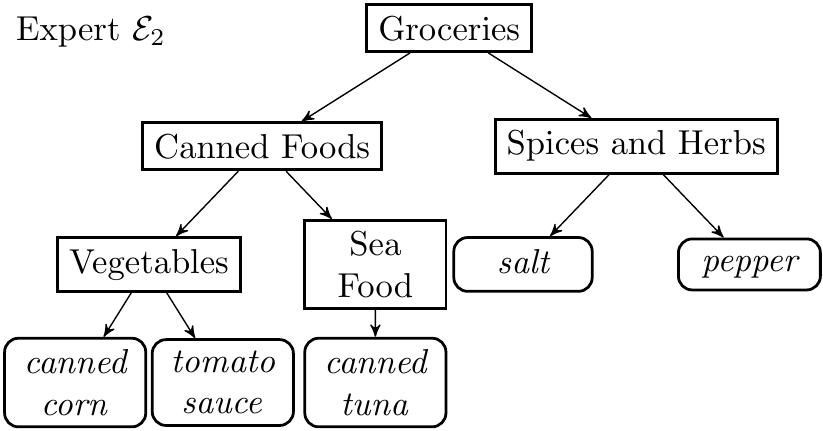}\hfill~
\caption{Two examples of expert hierarchies used to compute the semantic 
  similarities between object classes. For example, expert~$\mathcal{E}_1$ on 
    the left assigns a similarity $\rho$ of 0.4 to the pair $\mathit{\{canned\ 
      corn, canned\ tuna\}}$, whereas $\mathcal{E}_2$ on the right assigns a 
      similarity of 0.33 to the same pair, see~\eqref{eq:wup}.}
  \label{fig:groceryExperts}
\end{figure*}

\subsubsection{Inferring a New User's Preferences}
\label{sec:onlineLearning}
 After acquiring probes for the new user, we can now append her column to the 
 ratings matrix and learn her biases and factors vector along with those of all 
 object-pairs and other users in the system as in \eqref{eq:optimization}. In 
 practice, we can avoid re-learning the model for all users and object-pairs 
 known in the system. Note that the computation of the factorization will 
 require more time as the number of known ratings in $\ratingMat$ increases or 
 for higher values of $K$. Here, we propose a more efficient technique suitable 
 for inferring a new user's preferences given a previously learned 
 factorization.  After learning with all object-pairs and users in the database, 
 we assume that all object-pair biases $\biasPair$ and factor vectors 
 $\mathbf{S}$ are fixed, and can be used to model the preferences of new users.  
 We can then formulate a smaller problem to learn the bias $\biasUser$ and 
 factors vector $\mathbf{t}_j$ of the new user $u_j$ based on the probe ratings 
 we have for this user, i.e.,
\begin{equation}
\begin{aligned}
 \biasUser, \mathbf{t}_j &= \argmin_{\biasUser,\mathbf{t}_j} \sum_{i\in \mathcal{I}_j} 
 (e_{ij})^2 + \frac{\lambda}{2}(\biasUser^2 + \|\mathbf{t}_j\|^2),\\
 &= \argmin_{\biasUser,\mathbf{t}_j} \sum_{i\in \mathcal{I}_j} (r_{ij} - (\mu + \biasPair + 
     \biasUser + \mathbf{s}_i^T \cdot \mathbf{t}_j))^2 +\\
   &\ \ \ \ \ \frac{\lambda}{2}(\biasUser^2 + \|\mathbf{t}_j\|^2).
\end{aligned}
\label{eq:newUserLearning}
\end{equation}
Note that, in general, the inclusion of the ratings of a new user in
$\ratingMat$ will affect the biases and factor vectors of the object-pairs.  
Whereas \eqref{eq:optimization} represents the batch learning problem to update 
the model for all users and object-pairs, \eqref{eq:newUserLearning} assumes 
that the object-pair biases and factor vectors have already been learned from a 
sufficiently-large set of users that is representative of the new user. This can 
be useful  in a lifelong learning scenario where the robot can efficiently make 
predictions for a new user when solving a tidy-up task.  With more knowledge 
accumulated about the new users, we can update the factorization model and 
biases for all object-pairs and users in a batch manner.

\section{Mixture of Experts for Predicting Preferences of Unknown Objects}
\label{sec:wup}
Thus far, we presented how our approach can make predictions for object-pairs 
that are known to the robot. In this section, we introduce our approach for 
computing predictions for an object-pair that no user has rated before, for 
example when the robot is presented with an object $o_*$ that is not in 
$\objects$.  There, we cannot rely on standard collaborative filtering since we 
have not learned the similarity of the pair (through its factors vector) to 
others in $\pairs$.

Our idea is to leverage the known ratings in $\ratingMat$ as well as
prior information about object similarities that we mine from the internet. The 
latter consists of object hierarchies provided by popular websites, including 
online supermarkets, stores, dictionaries, etc. \figref{fig:groceryExperts} 
illustrates parts of two example experts for a grocery scenario.  Formally, 
            rather than relying on one source of information, we adopt a 
            \emph{mixture of experts} approach where each expert $\mathcal{E}_i$ 
            makes use of a mined hierarchy that provides information about 
            similarities between different objects. The idea is to query the 
            expert about the unknown object $o_*$ and retrieve all the 
            object-pair preferences related to it. The hierarchy is a graph or a 
            tree where a node is an object and an edge represents an ``is-a'' 
            relation.

When the robot is presented with a set of objects to organize that includes a 
new object $o_*$, we first ignore object-pairs involving $o_*$ and follow our 
standard collaborative filtering approach to estimate preferences for all other 
object-pairs, i.e., \eqref{eq:ratingDetailed}.  To make predictions for 
object-pairs related to the new object, we compute the similarity $\rho$ of 
$o_*$ to other objects using the hierarchy graph of the expert. For that, we 
employ the $\mathit{wup}$ similarity~\cite{wup94}, a measure between 0 and 1 
used to find semantic similarities between concepts
\begin{equation}
\label{eq:wup}
\rho_{lk} = 
\frac{\mathit{depth}(\mathit{LCA}(o_l,o_k))}{0.5(\mathit{depth}(o_l)+\mathit{depth}(o_k))},
\end{equation}
where $\mathit{depth}$ is the depth of a node, and $\mathit{LCA}$ denotes the 
lowest common ancestor. In the example of expert $\mathcal{E}_1$ in 
\figref{fig:groceryExperts}-left, the lowest common ancestor of \emph{canned 
  corn} and \emph{canned tuna} is Canned Foods. Their $\mathit{wup}$ similarity 
  based on $\mathcal{E}_1$ and $\mathcal{E}_2$ 
  (\figref{fig:groceryExperts}-right) is 0.4 and 0.33, respectively.  
  Note that in general, multiple paths could exist between two object 
    classes in the same expert hierarchy. For example, $\mathit{coffee}$ could 
      be listed under both Beverages and Breakfast Foods. In such cases, we take 
      the path ($\mathit{LCA}$) that results in the highest $\mathit{wup}$ 
      measure for the queried pair.

Given this similarity measure, our idea is to use the known ratings of objects 
similar to $o_*$ in order to predict the ratings related to it. For example, if 
$\mathit{salt}$ is the new object, we can predict a rating for $\mathit{\{salt, 
  coffee\}}$ by using the rating of $\mathit{\{pepper, coffee\}}$ and the 
  similarity of $\mathit{salt}$ to $\mathit{pepper}$. We compute the expert 
  rating $\hat{r}_{\mathcal{E}_i}(o_*, o_k)$ for the pair $\{o_*,o_k\}$ as the 
  sum of a baseline rating, taken as the similarity $\rho_{*k}$, and a weighted 
  mean of the residual ratings for similar pairs, i.e.,
\begin{equation}
\label{eq:expertPred}
\hat{r}_{\mathcal{E}_i}(o_*, o_k) = \rho_{*k} + \eta_1\sum_{l \in \mathcal{L}} 
\rho_{*l}  \, \,  (r(o_l, o_k) - \rho_{lk}),
\end{equation}
where $\eta_1 = 1/\sum_{l \in \mathcal{L}} \rho_{*l}$ is a normalizer, and 
$\mathcal{L}$ is the set of object indices such that the user's rating of pair 
$\{o_l,o_k\}$ is known. In other words, we rely on previous preferences of the 
user ($r(o_l, o_k)$) combined with the similarity measure extracted from the 
expert. The expert hierarchy captures one strategy for organizing the objects by 
their similarity. If this perfectly matches the preferences of the user, then 
the sum in \eqref{eq:expertPred} will be zero, and we simply take the expert's 
baseline $\rho_{*k}$ when predicting the missing rating. Otherwise, we correct 
the baseline based on how much the similarity measure deviates from the known 
ratings of the user.

Accordingly, each of our experts predicts a rating using its associated  
hierarchy.  We compute a final prediction $\hat{r}_{\mathcal{E}_*}$ as a 
combined estimate of all the expert ratings:
\begin{equation}
\label{eq:expertsMerged}
\hat{r}_{\mathcal{E}_*}(o_*, o_k) = \eta_2\sum_i w_i \, \hat{r}_{\mathcal{E}_i} 
(o_*, o_k),
\end{equation}
where $w_i \in [0,1]$ represents the confidence of $\mathcal{E}_i$,
$\mathcal{E}_*$ denotes the mixture of experts, and $\eta_2 = 1/\sum_i
w_i$ is a normalizer. We compute the confidence of expert 
  $\mathcal{E}_i$ as $w_i = \exp(-e_i)$, where $e_i$ is the mean error in the 
  expert predictions when performing a leave-one-out cross-validation on the 
  known ratings of the user as in \eqref{eq:expertPred}. We set this score to 
  zero if it is below a threshold, which we empirically set to 0.6 in our work.  
  Moreover, we disregard the rating of an expert if $o_*$ cannot be found in its 
  hierarchy, or if all associated similarities $\rho_{*l}$ to any relevant 
  object $o_l$ are smaller than 0.4.
   
   Note that in general, both objects in a new pair could have been previously 
   encountered by the robot separately, but no rating is known for them 
   together. When retrieving similar pairs to the new object-pair, we consider 
   the similarities of both objects in the pair to other objects.  For example, 
   we can predict the rating of $\{\mathit{sugar}, \mathit{coffee}\}$ by 
   considering the ratings of both $\{\mathit{flour}, \mathit{coffee}\}$ and 
   $\{\mathit{sugar}, \mathit{tea}\}$.

\section{Grouping Objects Based on Predicted Preferences}
\label{sec:spectralClustering}
\begin{figure}[t]
 \centering
\vspace{2mm}
\includegraphics[width=0.9\columnwidth]{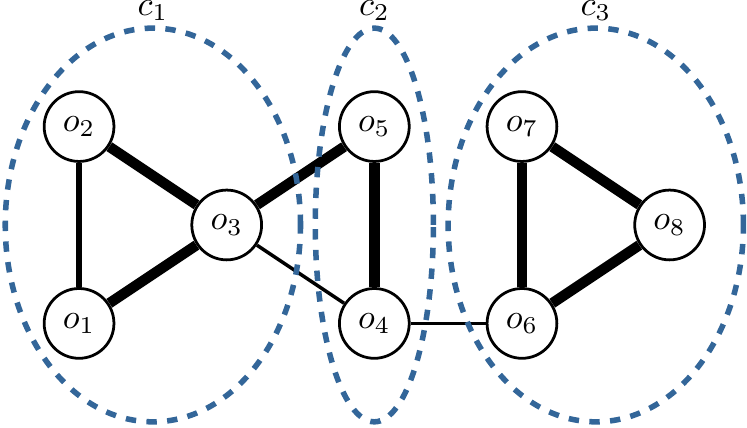}\\
\vspace{5mm}
\includegraphics[height=3.2cm]{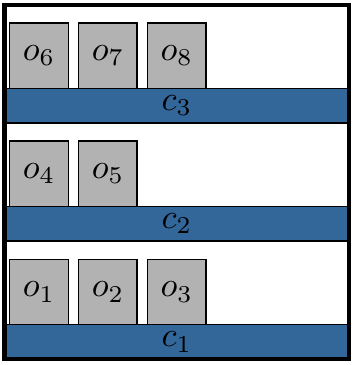}
\caption{Top: a graph depicting the relations between objects.  Each node 
  corresponds to an object, and the weights (different edge thickness) 
    correspond to the pairwise ratings. We partition the graph into subgraphs 
    using spectral clustering. Bottom: we assign objects in the same subgraph to 
    the same container.}
\label{fig:graph}
\end{figure}

Now that it is possible to compute pairwise object preferences about
known or unknown objects, we aim to sort the objects into different
containers. In general, finding a partitioning of objects such that
all pairwise constraints are satisfied is a non-trivial task. For
example, the user can have a high preference for $\mathit{\{pasta,
  rice\}}$ and for $\mathit{\{pasta, tomato\ sauce\}}$, but a low
preference for $\mathit{\{rice, tomato\ sauce\}}$.  Therefore, we aim
at satisfying as many of the preference constraints as possible when
grouping the objects into $C'\leq C$ containers, where $C$ is the
total number of containers the robot can use.

First, we construct a weighted graph where the nodes represent the
objects, and each edge weight is the rating of the corresponding
object-pair, see~\figref{fig:graph}. We formulate the subdivision of objects 
into $C'$ containers as a problem of partitioning of the graph into
$C'$ subgraphs such that the cut (the sum of the weights between the
subgraphs) over all pairs of subgraphs is minimized. This is called
the minimum $k$-cut problem~\cite{minKCut94}. Unfortunately, finding
the optimal partitioning of the graph into $C'\leq C$ subgraphs is
NP-hard.  In practice, we efficiently solve this problem by using a
spectral clustering approach~\cite{spectral96}. The main idea is to
partition the graph based on the eigenvectors of its Laplacian matrix,
$L$, as this captures the underlying connectivity of the graph.

Let $V$ be the matrix whose columns are the first $C'$ eigenvectors of
$L$. We represent each object by a row of the matrix $V$, i.e., a
$C'$-dimensional point, and apply $k$-means clustering using $C'$
clusters to get a final partitioning of the objects. To estimate the best number 
of clusters, we implement a self-tuning heuristic which sets the number of 
clusters $C'$ based on the location of the biggest eigen-gap from the 
decomposition of $L$, which typically indicates a reliable way to partition the 
graph based on the similarities of its nodes. A good estimate for this is the 
number of eigenvalues of $L$ that are approximately 
zero~\cite{luxburgTutorial07,zelnik2004self}. If there exist less containers in 
the environment than this estimate, we use all $C$ containers for partitioning 
the objects.

\begin{figure*}[t]
\setlength{\fboxsep}{0pt}
\centering
~\hfill\fbox{\includegraphics[height=4cm]{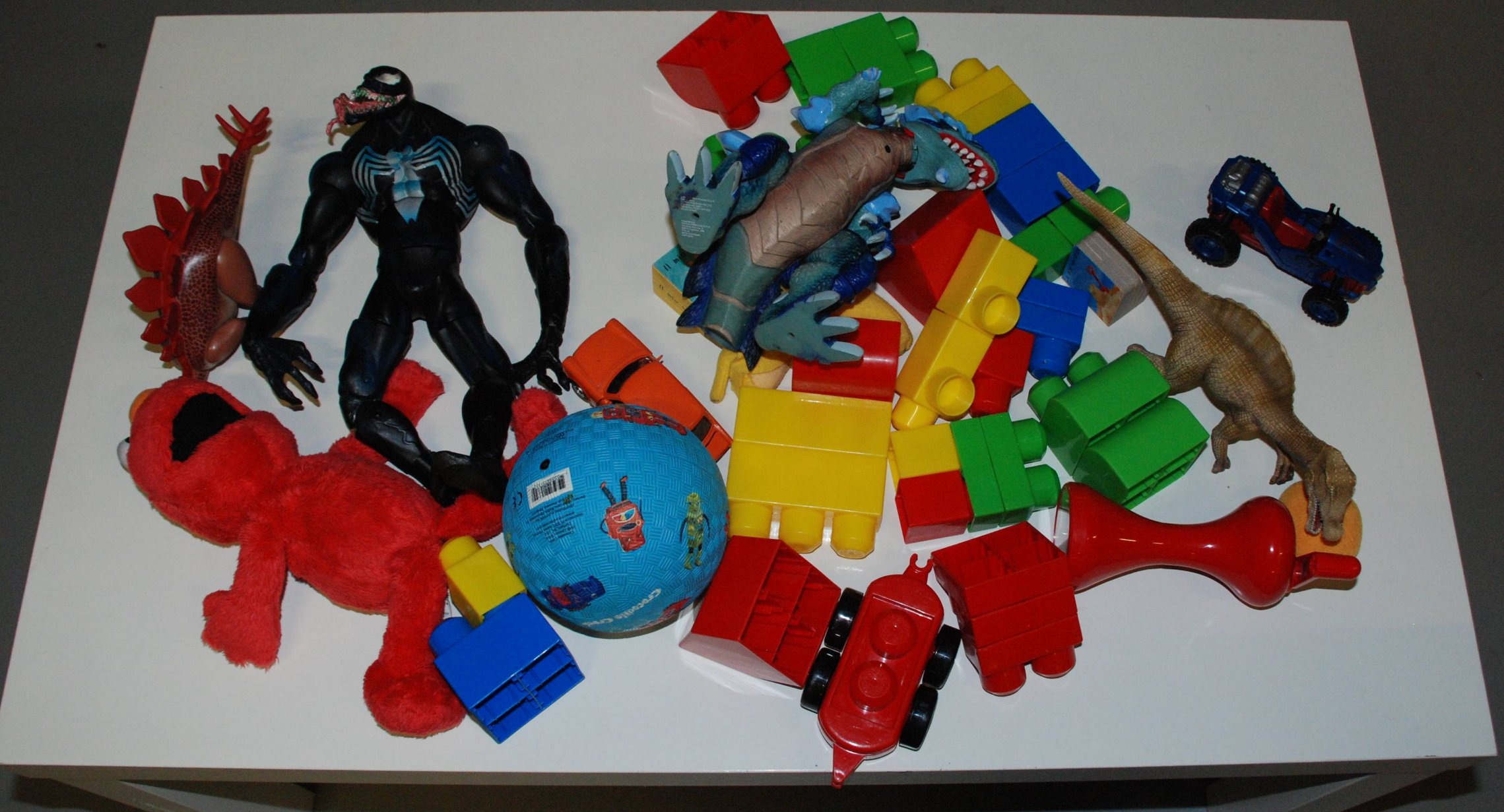}}\hfill
\includegraphics[height=4cm]{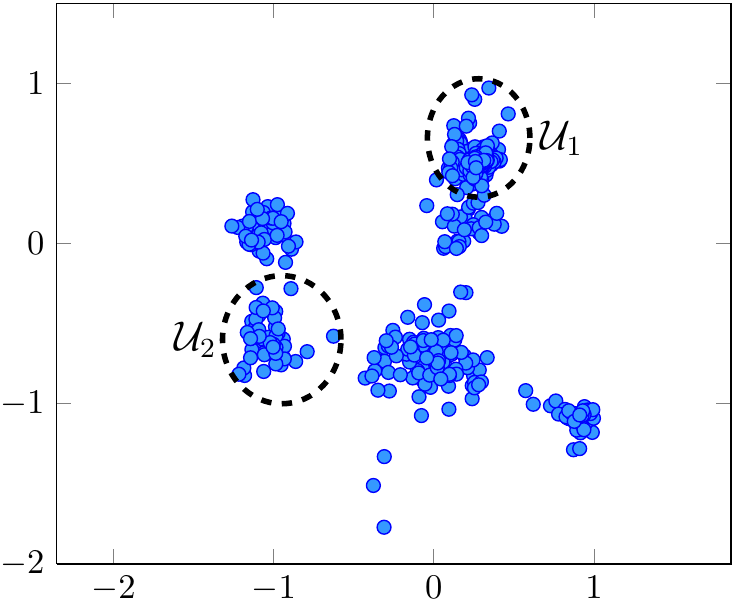}\hfill~
\caption{Left: we considered a scenario of organizing toys in boxes. Right: a 
  visualization of user tastes with respect to organizing toys, where we plot 
    the user factor vectors projected to the first two dimensions.  For example, 
        the cluster $\users_1$ corresponds to users who grouped all building 
          blocks together in one box.  Cluster $\users_2$ corresponds to users 
          who separated building blocks into standard bricks, car-shaped blocks,          
        and miscellaneous.}
\label{fig:toys}
\end{figure*}

\section{Experimental Evaluation}
\label{sec:experiments}
In this section, we present the experimental evaluation of our approach by 
testing it on two tidy-up scenarios. We first demonstrate different aspects of 
our approach for a simple scenario of organizing toys in boxes based on a small 
dataset with 15 survey participants. In the second scenario, we address sorting 
grocery items on shelves, and provide an extensive evaluation based on ratings 
we collected from over 1,200 users using crowdsourcing.
We demonstrate that:
\emph{i}) users indeed have different preferences with respect to sorting 
objects when tidying up, \emph{ii}) our approach can accurately predict personal 
user preferences for organizing objects (\secref{sec:cfLearning}), \emph{iii}) 
  we are able to efficiently and accurately learn a model for a new user's 
  preferences based on previous training users~(\secref{sec:cfNewUsers}), 
  \emph{iv}) our mixture of experts approach enables making reasonable 
  predictions for previously unknown objects (\secref{sec:wup}), \emph{v}) our 
  approach is suitable for life-long learning of user preferences, improving 
  with more knowledge about different users, \emph{vi}) our object partitioning 
  approach based on spectral clustering can handle conflicting pairwise 
  preferences and is flexible with respect to the number of available containers 
  (\secref{sec:spectralClustering}), and \emph{vii}) our approach is applicable 
  on a real tidy-up robot scenario.

In the following experiments, we evaluate our approach using two different 
methods for acquiring probe ratings, and compare our results to different 
baselines. For that, we use the following notation:
\begin{itemize}
\item CF refers to our collaborative filtering approach for learning user 
preferences, as described in \secref{sec:cfLearning}. When selecting probes to 
learn for a new user, we do so by clustering the object-pairs based on their 
learned factor vectors in order to query the user for a range of preferences, 
        see~\secref{sec:probing}.
\item CF-rand selects probes randomly when learning for a new user and then uses 
our collaborative filtering approach to make predictions as in 
\secref{sec:cfLearning}.
\item CF-rand$'$ selects probes randomly and learns the preferences of a new 
user based on the object-pair biases and factor vectors learned from previous 
users as in~\secref{sec:onlineLearning}.
\item Baseline-I uses our probing approach as in CF, and then predicts each 
unknown pair rating as the mean rating over all users who rated it.
\item Baseline-II selects probes randomly and then predicts each
unknown pair rating as the mean rating over all users.
\end{itemize}
In all experiments, unless stated otherwise, we set the number of factor 
dimensions to $K=3$ and the regularizer to~$\lambda~=~0.01$. As part of 
  our implementation of \eqref{eq:optimization} and \eqref{eq:newUserLearning}, 
  we rely on the L-BFGS implementation by     
  \citeauthor{liblbfgs}~\cite{liblbfgs}. Note that in our work, we assume that 
  the robot is equipped with suitable techniques for recognizing the objects of 
  interest. In our experiments, we relied on fiducial markers attached to the 
  objects, and also implemented a classifier that recognizes grocery items by 
  matching visual features extracted from the scene to a database of product 
  images.

\subsection{Task 1: Organizing Toys}
\label{sec:expToy}
In this experiment, we asked 15 people to sort 26 different toys in
boxes, see~\figref{fig:toys}-left. This included some plush
toys, action figures, a ball, cars, a flashlight, books, as well as different
building blocks.  Each participant could use \emph{up to} six boxes to
sort the toys. Overall, four people used four boxes, seven people used
five boxes, and four people used all six available boxes to sort
the toys.

We collected these results in a ratings matrix with 15 user columns
and 325 rows representing all pairs of toys. Each entry in a user's
column is based on whether the user placed the corresponding objects in
the same box or not, see~\secref{sec:probing}. For a fine quantification, we 
used these ratings to bootstrap a larger ratings matrix representing a noisy 
version of the preferences with 750 users. For this, we randomly selected 78 
ratings out of 325 from each column. We repeated this operation 50 times for 
each user and constructed a ratings matrix of size 325$\times$750 where 76$\%$ 
of the ratings are missing.

As a first test, we computed a factorization of the ratings matrix as described
in~\secref{sec:cfLearning}.  \figref{fig:toys}-right shows the user factors 
$\mathbf{T}$ projected to the first two dimensions, giving a visualization of 
the user tastes. For example, the cluster of factors labeled $\users_1$ 
corresponds to users who grouped all building blocks together in one box.

\subsubsection{Predicting User Preferences for Pairs of Toys}
\label{sec:toysFactors}
\begin{figure}[t]
\centering
\includegraphics[width=0.8\columnwidth]{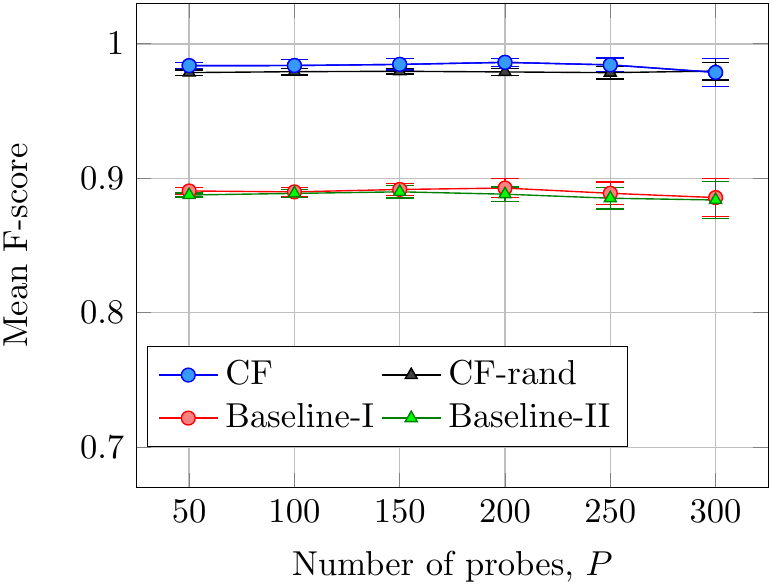}\\[5mm]
\includegraphics[width=0.8\columnwidth]{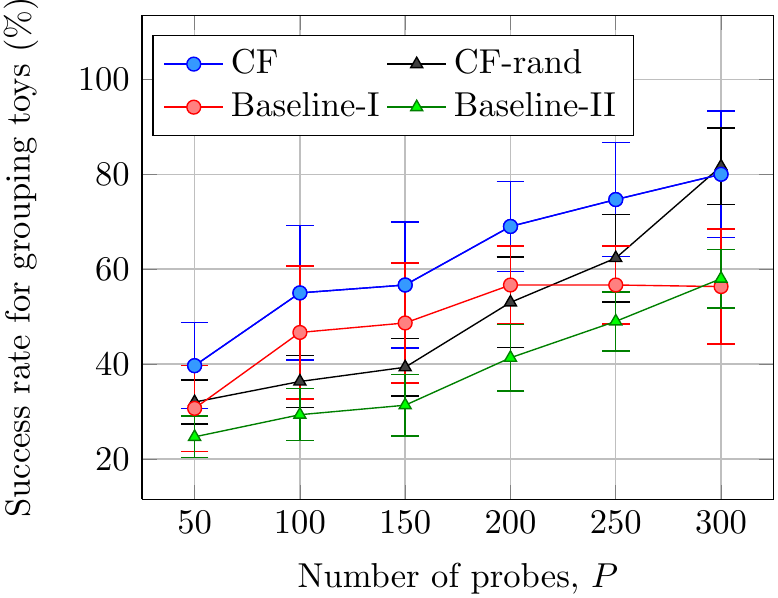}
\caption{Top: the mean F-score of the predictions of our
  approach (CF) in the toys scenario for different numbers of known
  probe ratings. We achieve an F-score of 0.98-0.99 on average over
  all predicted ratings.  CF-rand selects probes randomly and then
  uses our approach for predicting. It is able to achieve an F-score of 0.98. On 
  the other hand, baselines I and II are unable to adapt to multimodal user 
  preferences.  Bottom: the percentage of times our approach is able to predict 
  the correct arrangement of boxes for sorting different toys. We outperform 
  both baselines and improve with more probe ratings as expected, reaching a 
  success rate of 80$\%$. By selecting probes based on object-pair factor 
  vectors, we are able to achieve higher success rates with less probes compared 
  to CF-rand.}
\label{fig:resultsToys}
\end{figure}

We evaluated our approach for predicting the preferences of the 15
participants by using the partial ratings in the matrix we constructed
above.  For each of the participants, we queried for the ratings of
$P$ probes. We hid all other ratings from the user's column and predicted them 
using the ratings matrix and our approach. We rounded each prediction to the 
nearest integer on the rating scale [0,1] and compared it to the ground truth
ratings. We evaluated our results by computing the precision, recall,
and F-score of our predictions with respect to the two rating classes:
\emph{no} ($r=0$), and \emph{yes} ($r=1$). We set the number of probes to $P$ = 
50, 100, \dots, 300 known ratings, and repeated the experiment 20 times for each 
value, selecting different probes in each run. The mean F-scores of both rating
classes, averaged over all runs are shown in~\figref{fig:resultsToys}-top.

Both collaborative filtering techniques outperform baselines I and II. On 
average, CF and CF-rand maintain an F-score around 0.98 over all predicted pair 
ratings. On the other hand, Baseline-I and Baseline-II achieve an F-score of 
0.89 on average. By employing the same strategy for all users, these baselines 
  are only able to make good predictions for object-pairs that have a unimodal 
  rating distribution over all users, and cannot generalize to multiple tastes 
  for the same object-pair.

\subsubsection{Sorting Toys into Boxes}
We evaluated our approach for grouping toys into different boxes based on the 
predicted ratings in the previous experiment. For each user, we partitioned the 
objects into boxes based on the probed and predicted ratings as described 
in~\secref{sec:spectralClustering}, and compared that
to the original arrangement. We computed the success rate, i.e.,
the percentage of cases where we achieve the same number and content of boxes,
see~\figref{fig:resultsToys}-bottom.  Our approach has a success rate
of 80$\%$ at $P=300$.  As expected, the performance improves with the
number of known probe ratings.  On the other hand, even with $P=300$
known ratings, Baseline-I and Baseline-II have a success rate of only
56$\%$ and 58$\%$.  Whereas CF-rand achieves a success rate of 82$\%$
at $P=300$, it requires at least 200 known probe ratings on average to
achieve success over 50$\%$.  On the other hand, CF achieves a success rate of
55$\%$ with only 100 known probe ratings. The probes chosen by our
approach capture a more useful range of object-pairs based on the distribution 
of their factor vectors, which is precious information to distinguish a user's 
taste.

\subsubsection{Predicting Preferences for New Objects}
\label{sec:toysRUs}
We evaluated the ability of our approach to make predictions for
object-pairs that no user has rated before~(\secref{sec:wup}). For
each of the 26 toys, we removed all ratings related to that toy from
the ratings of the 15 participants.  We predicted those pairs using a
mixture of three experts and the known ratings for the remaining
toys. We evaluated the F-scores of our predictions as before by
averaging over both \emph{no} and \emph{yes} ratings. We based our
experts on the hierarchy of an online toy store (toysrus.com),
appended with three different hierarchies for sorting the building
blocks (by size, color, or function).
The expert hierarchies contained between 165-178 nodes. For one of the
toys (flash light), our approach failed to make predictions since the
experts found no similarities to other toys in their hierarchy. For
all other toys, we achieved an average F-score of 0.91 and predicted
the correct box to place a new toy 83$\%$ of the time.

\begin{figure*}
\centering
\includegraphics[width=0.68\textwidth]{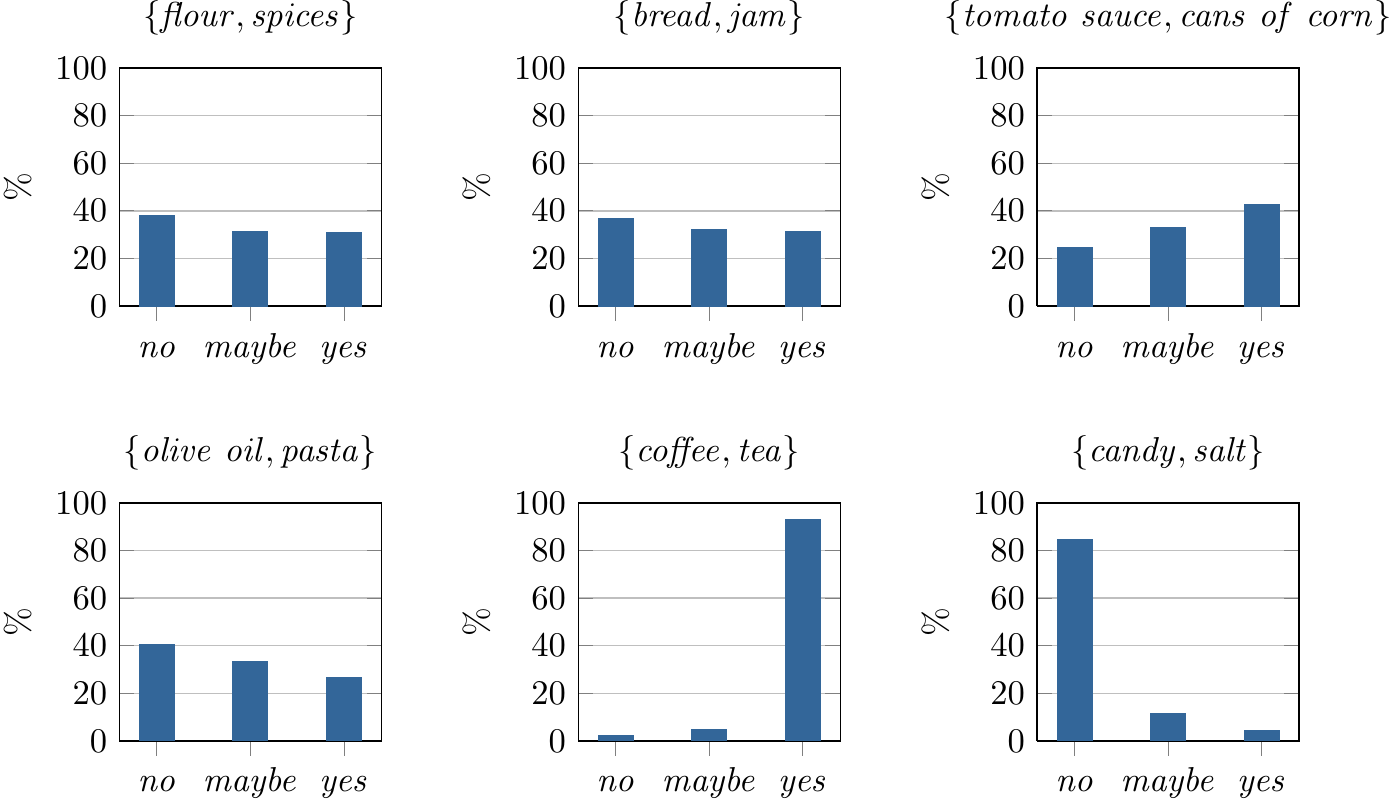}
\caption{Example distributions of the ratings given by users for different 
  object-pairs. Each user could answer with \emph{no} ($r=0$), \emph{maybe} 
  ($r=0.5$), or \emph{yes} ($r=1$) to indicate the preference for placing the 
    two objects on the same shelf. The three possible rating classes, as well as 
    the noise inherent to crowdsourcing surveys, resulted in multi-modal taste 
    distributions. This highlights the difficulty of manually designing rules to 
    guide the robot when sorting objects into different containers.}
\label{fig:ratingExamplesGroceries}
\end{figure*}

\subsection{Task 2: Organizing Groceries}
\label{sec:groceries}
In this scenario, we considered the problem of organizing different grocery 
items on shelves. We collected data from over 1,200 users using a crowdsourcing 
service~\cite{crowdFlower}, where we considered a set of 
22 common grocery item types, e.g., cans of beans, flour, tea, etc.
We asked each user about her preferences for a subset of pairs related
to these objects. For each pair, we asked the user if she would
place the two objects together on the same shelf. Each user could
answer with \emph{no}, \emph{maybe}, or \emph{yes}, which we
translated to ratings of 0, 0.5, and 1, respectively. We aggregated the answers 
into a ratings matrix $\ratingMat$ of size 179$\times$1,284. Each of the user 
columns contains between 28 and 36 known ratings, and each of the 179 
object-pairs was rated between 81 to 526 times. Overall, only around $16\%$ of 
the matrix is filled with ratings, with the ratings distributed as in 
\tabref{tab:NoMaybeYes}. Due to the three possible ratings and the noise 
inherent to crowdsourcing surveys, the ratings we obtained were largely 
multi-modal, see~\figref{fig:ratingExamplesGroceries} for some examples.

\subsubsection{Predicting User Preferences for Pairs of Grocery Items}
\label{sec:GroceriesProbing}

\begin{figure}
\centering
\includegraphics[width=0.7\columnwidth]{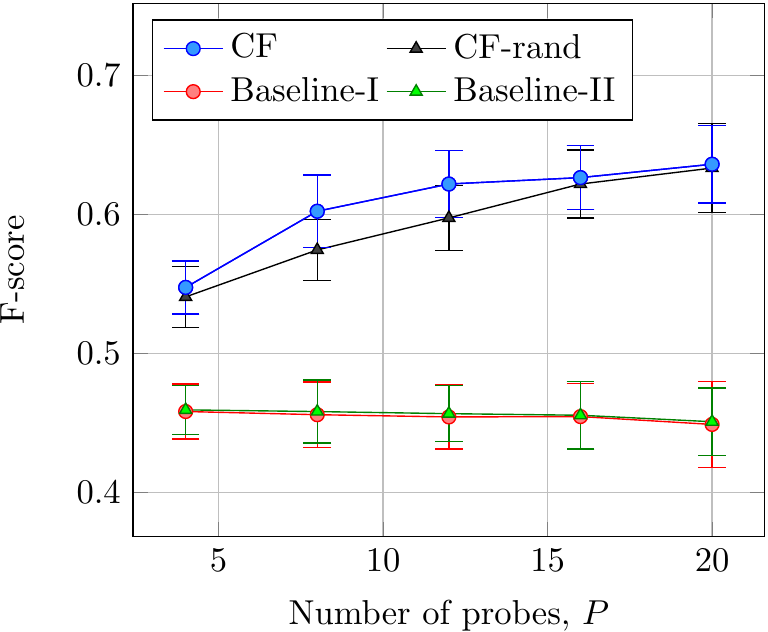}\\[5mm]
\includegraphics[width=0.7\columnwidth]{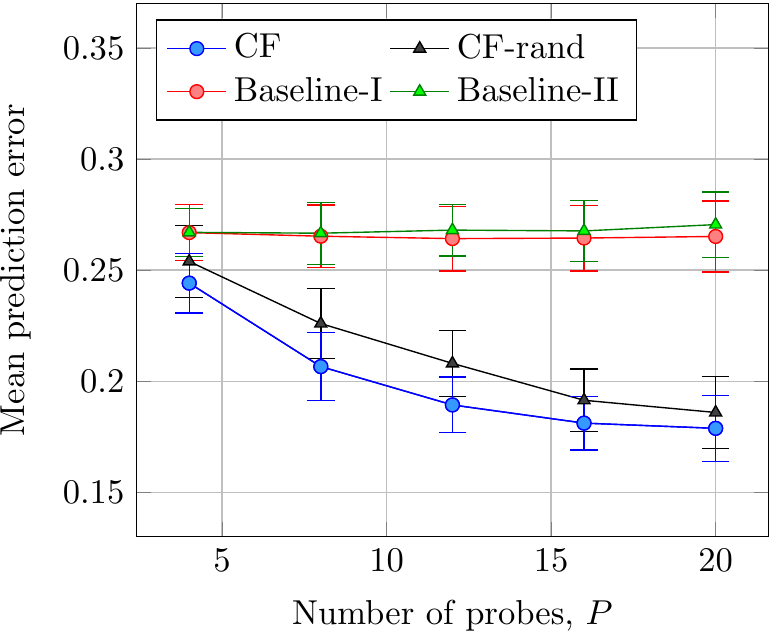}
                       \caption{Results for the scenario of organizing
                         grocery items on different shelves. Top: the
                         mean F-score of our predictions averaged over
                         all rating classes \emph{no}, \emph{maybe},
                         and \emph{yes}. Despite the large degree of
                         multi-modality and noise in the user
                         preferences we collected through
                         crowdsourcing, our approach (CF) is able to
                         achieve an F-score of 0.63 with 20 known
                         probes and to outperform the
                         baselines. Moreover, our performance improves
                         with more knowledge about user preferences as
                         expected. Bottom: the mean prediction error for 
                         different numbers of probes, $P$. The baselines are 
                         unable to cope with different modes of user 
                         preferences. They consistently result in a prediction 
                         error of around $0.27$ irrespective of the number of 
                         probes. On the other hand, the mean prediction error 
                         using CF $0.24$ to $0.18$ as $P$ increases from 
            4 to 20. Using our probing technique, we are able to achieve a lower 
              error with fewer probes compared to CF-rand.}
\label{fig:resultsGroceries}
\end{figure}

\begin{figure}[t]
\centering
{
\footnotesize
\renewcommand{\arraystretch}{1.5}
\begin{tabular}[b]{cc|c|c|c|}
\cline{3-5}
& &\emph{no}&\emph{maybe} &\emph{yes}\\  \cline{3-5}
\cline{1-5}
\multicolumn{1}{ |c }{\multirow{3}{*}{Baseline-I} } &
\multicolumn{1}{ |c| }{Precision} & $0.71$ & $0.34$ &$0.79$    \\ \cline{2-5}
\multicolumn{1}{ |c  }{}                        &
\multicolumn{1}{ |c| }{Recall} & $0.52$ & $0.69$ & $0.19$ \\ \cline{2-5}
\multicolumn{1}{ |c  }{}                        &
\multicolumn{1}{ |c| }{F-score} & $0.60$ & $0.46$ & $0.31$ \\ \cline{1-5}
\multicolumn{1}{ |c  }{\multirow{3}{*}{CF}  } &
\multicolumn{1}{ |c| }{Precision} & $0.80$ & $0.45$ & $0.72$  \\ \cline{2-5}
\multicolumn{1}{ |c  }{}                        &
\multicolumn{1}{ |c| }{Recall} & $0.72$ & $0.63$ & $0.49$  \\ \cline{2-5}
\multicolumn{1}{ |c  }{}                        &
\multicolumn{1}{ |c| }{F-score} & $0.76$ & $0.53$ & $0.58$  \\ \cline{1-5}
\end{tabular}}\\[5mm]
\includegraphics[width=0.72\columnwidth]{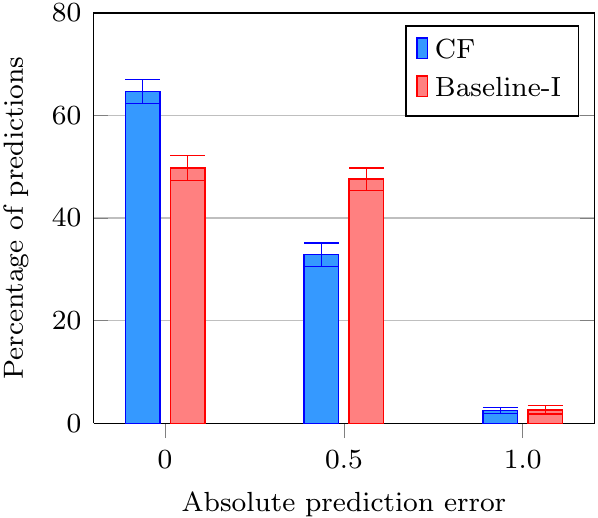}
\caption{Top: the detailed evaluation for the groceries scenario with $P=12$ 
  probes.  Our approach results in higher F-scores across all rating classes 
    compared to the baseline. \figref{fig:resultsGroceries}-top shows the mean 
    F-score for different values of $P$. Bottom: the detailed distribution of 
    the prediction errors using $P=12$ probes,                
        see~\figref{fig:resultsGroceries}-bottom for the mean error for 
          different values of $P$.}
\label{fig:groceriesP12}
\end{figure}

\begin{table}
\centering
\normalsize
\caption{The distribution of ratings for the groceries scenario obtained through 
  crowdsourcing. Overall, we gathered 37,597 ratings about 179 object-pairs from 
    1,284 users. For each object-pair, users indicated whether they would place 
    the two objects on the same shelf or not.}
\label{tab:NoMaybeYes}
\begin{tabular}{c|c|c|c|}
\cline{2-4} & \multicolumn{3}{ c| }{Rating Classes} \\
\cline{2-4} & \emph{no} & \emph{maybe} &  \emph{yes}\\
            & ($r=0$) & ($r=0.5$)& ($r=1$)\\
\cline{1-4} \multicolumn{1}{ |c| } {Rating Percentage} & $47.9\%$ & $29.2\%$ & 
$22.9\%$ \\
\cline{1-4}
\end{tabular}
\end{table}
We show that our approach is able to accurately predict user ratings of 
object-pairs using the data we gathered from crowdsourcing.  For this, we tested 
our approach through 50 runs of cross-validation.  In each run, we selected 50 
user columns from $\ratingMat$ uniformly at random, and queried them with $P$ of 
their known ratings.  We hid the remaining ratings from the matrix and predicted 
them using our approach. We rounded each prediction to the closest rating 
(\emph{no}, \emph{maybe}, \emph{yes}) and evaluated our results by
computing the precision, recall, and F-score. Additionally, we
compared the predictions of our approach (CF) to CF-rand, Baseline-I,
and Baseline-II described above. The average F-scores
over all runs and rating classes are shown in~\figref{fig:resultsGroceries}-top 
for $P=$ 4, 8, \dots, 20. Both collaborative filtering approaches outperform the 
baseline approaches, reaching a mean F-score of 0.63 at $P=$ 20 known probe
ratings. Baseline-I and Baseline-II are only able to achieve an
F-score of 0.45 by using the same rating of a pair for all users. Note
that by employing our probing strategy, our technique is able to
achieve an F-score of 0.6 with only 8 known probe ratings. On the
other hand, CF-rand needs to query a user for the ratings of at least 12 
object-pairs on average to achieve the same performance.
 For a closer look at the performance with respect to the three rating classes, 
 we select the results at $P=12$ and show the per-class precision, recall, and 
 F-score values for both CF and Baseline-I in \figref{fig:groceriesP12}-top.  
 Note that the baseline achieves its highest recall value for the \emph{maybe} 
 class since it uses the mean rating received by a specific object-pair to 
 predict its rating for new users.  On the other hand, we are able to achieve a 
 similar recall (0.63) for the \emph{maybe} class, as well as higher recall 
 values for the \emph{no} and \emph{yes} classes despite the large degree of 
 noise and the variance in people preferences in the training data.  Our 
 approach is able to achieve higher F-scores over all rating classes compared to 
 the baseline.  Out of the three classes, we typically achieved better scores 
 for predicting the \emph{no} class compared to \emph{maybe} or \emph{yes}. This 
 is expected due to the distribution of the training ratings we gathered from 
 the crowdsourcing data, see~\tabref{tab:NoMaybeYes}.

Additionally, we computed the prediction error (\eqref{eq:predError}) averaged 
over all experimental runs for each value of $P$, 
     see~\figref{fig:resultsGroceries}-bottom. The baselines are unable to cope 
     with the different modes of user preferences, and consistently result in a 
     prediction error of around $0.27$ irrespective of the number of probes. On 
     the other hand, the mean prediction error using CF and CF-rand drops from 
     $0.24$ to $0.18$ and from $0.25$ to $0.19$ as $P$ increases from 
            4 to 20, respectively. Note that, using our probing technique, we 
              are able to achieve a lower error with fewer probes compared to 
              CF-rand.  This illustrates the importance of selecting more 
              intelligent queries for users to learn their preferences. For a 
              closer inspection of the prediction error, 
              \figref{fig:groceriesP12}-bottom shows the distribution of the 
              error for our approach and Baseline-I given $P=12$ probes. Our 
              approach achieves an error of $0$ for $64.62\%$ of the predictions 
              we make, compared to $49.78\%$ only for Baseline-I. Moreover, 
              Baseline-I results in an absolute error of $0.5$ (confusing 
                  \emph{no}/\emph{yes} with \emph{maybe}) for $47.60\%$ of the 
              predictions, compared to $32.88\%$ only for our approach.  
              Finally, our approach and the baseline result in a prediction 
              error of $1.0$ (misclassifying \emph{no} as \emph{yes} or vice 
                  versa) for only $2.49\%$ and $2.62\%$ of the predictions, 
              respectively.
             
\subsubsection{The Effect of the Number of Latent Dimensions}
\label{sec:varyingNumFactors}
\begin{figure}[t]
\centering
\includegraphics[width=0.7\columnwidth]{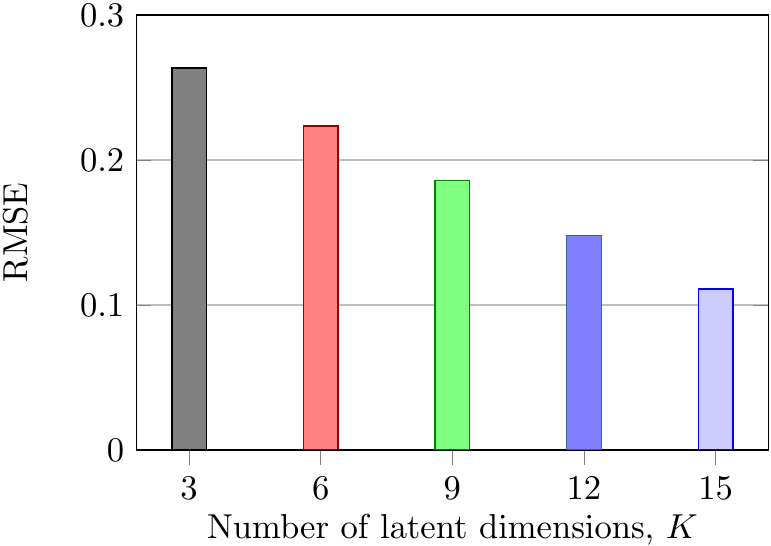}\\[5mm] 
\includegraphics[width=0.7\columnwidth]{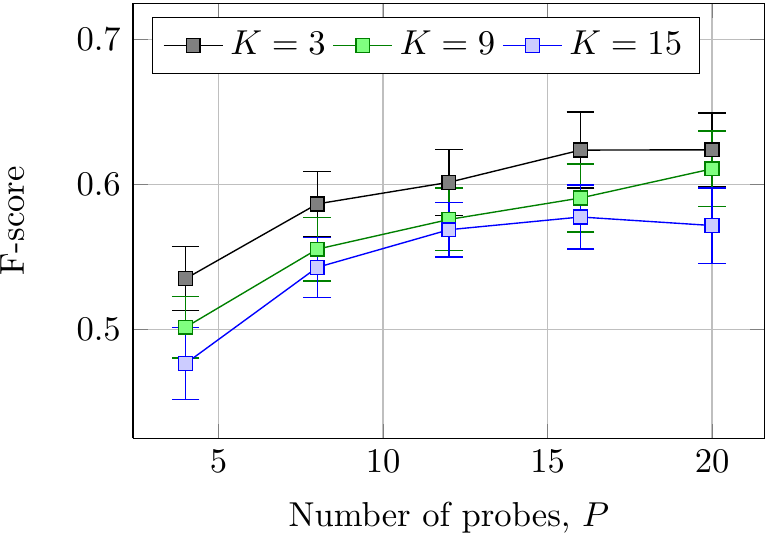}
                       \caption{We learned different factorizations of the 
                         ratings matrix $\ratingMat$ by varying the number of 
                           latent dimensions, $K$. For each learned model, we 
                           evaluated the RMSE when reconstructing the known 
                           ratings in $\ratingMat$ (top), and the F-score for 
                           predicting unknown ratings in $\ratingMat$ given 
                           different numbers of probes $P$ for randomly selected 
                           user columns (bottom). Learning factors 
                           ($\mathbf{S}                            $ and 
                            $\mathbf{T}$) with larger dimensionality leads to 
                           reconstructing the known ratings in $\ratingMat$ with 
                           a higher fidelity (lower RMSE).  However, this comes 
                           at the expense of over-fitting to the known ratings 
                           for users, leading to lower F-scores with larger $K$ 
                             when predicting new ratings given the same number 
                               of probes, $P$.}
\label{fig:resultsGroceriesRMSE}
\end{figure}

In this experiment, we investigated the effect of varying the number of latent 
dimensions $K$ used when learning the factorization of $\ratingMat$ on the 
quality of the learned model. We repeated the experiment in 
\secref{sec:GroceriesProbing} for $K= 3, 6, 9, 15$. For each setting of $K$, we 
conducted 50 runs where, in each run, we selected 50 random user columns, 
          queried them for $P$ random probe ratings, and learned the 
          factorization in \secref{sec:cfLearning} to predict the remaining 
          ratings. As in the previous experiment, we evaluated the quality of 
          predicting the unknown ratings by computing the average F-score for 
          the \emph{no}, \emph{maybe}, and \emph{yes} classes.  Additionally, we 
          computed the root mean square error (RMSE) for reconstructing the 
          \emph{known} ratings in $\ratingMat$ used in training, i.e.,

\begin{equation*}
\label{eq:rmse}
\text{RMSE} = \sqrt{\frac{1}{R} \sum_{i}\sum_{j\in\mathcal{J}_i}\Big(r_{ij} 
    - (\mu + \biasPair + \biasUser + \mathbf{s}_i^T \cdot \mathbf{t}_j)\Big)^2}.
\end{equation*}
 The results are shown in~\figref{fig:resultsGroceriesRMSE}. When using 
  larger values of $K$, we are able to reconstruct the known ratings in 
    $\ratingMat$ with lower RMSE values.  This is expected since we are 
    computing a more accurate approximation of $\ratingMat$ when factorizing it 
    into higher-dimensional matrices ($\mathbf{S}$ and $\mathbf{T}$), thus 
    capturing finer details in user preferences.  However, this results in 
    over-fitting (lower F-scores) when predicting unknown ratings, especially 
    for lower values of $P$. In other words, for higher values of $K$, we need 
      more probes per user when predicting unknown ratings, since we need to 
        learn more factors for each user and object-pair. In general, the more 
        known ratings we have in the user columns, the more sophisticated are 
        the models that we can afford to learn.

Furthermore, we found interesting similarities between object-pairs when 
inspecting their learned biases and factor vectors. For example (for $K=3$), 
           users tend to rate $\mathit{\{coffee, honey\}}$ similarly to 
           $\mathit{\{tea, sugar\}}$ based on the similarity of their factor 
           vectors. Also, the closest pairs to $\{\mathit{pasta},
\mathit{tomato\ sauce}\}$ included $\mathit{\{ pancakes, maple\
  syrup\}}$ and $\mathit{\{cereal, honey\}}$, suggesting that people often 
  consider whether objects can be used together or not.  With respect to the 
  biases ($\biasPair$) learned, object-pairs with the largest biases (rated above 
      average) included $\mathit{\{pepper, spices\}}$, $\mathit{\{pasta, 
    rice\}}$ and $\mathit{\{cans\ of\ corn, cans\ of\ beans\}}$.  Examples of 
    object-pairs with the lowest biases (rated below average) included 
    $\mathit{\{candy, olive\ oil\}}$, $\mathit{\{cereal, vinegar\}}$, and 
    $\mathit{\{cans\ of\ beans, cereals\}}$.  On the other hand, object-pairs 
    like $\mathit{\{cans\ of\ corn, pasta\}}$ and $\mathit{\{pancakes, honey\}}$ 
    had a bias of almost 0.

  \begin{figure}
   \centering
    \includegraphics[height=4cm]{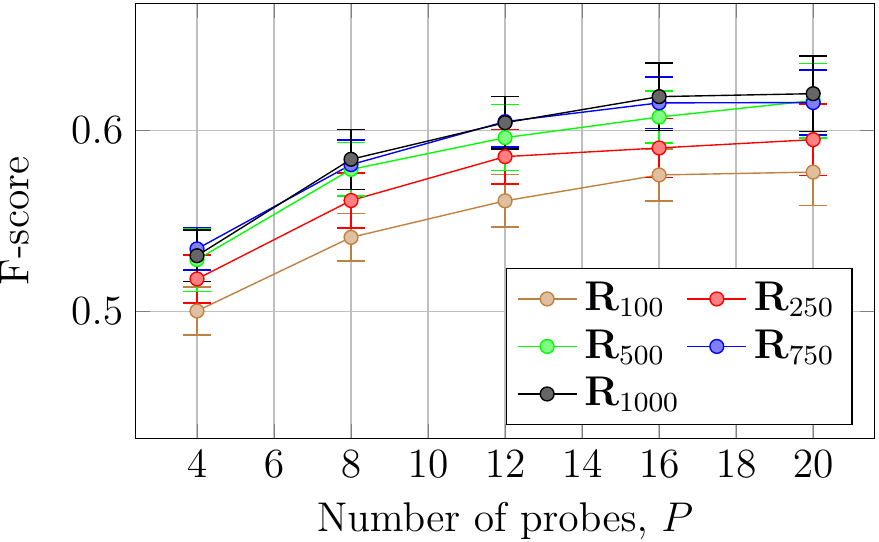}\\[5mm]
    \includegraphics[height=4cm]{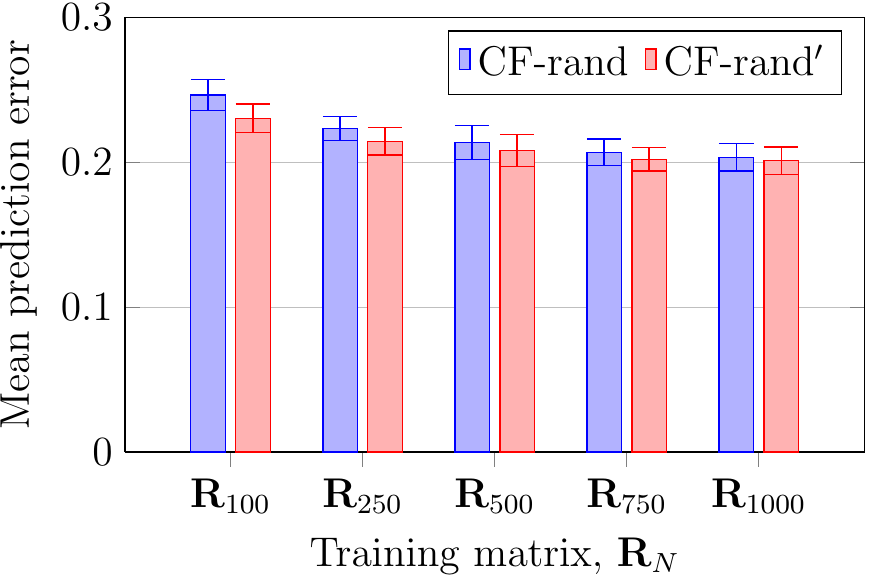}
  \caption{Top: we tested our approach for learning the preferences of 
    new users based on the object-pair biases and factors vectors learned from 
      rating matrices $\ratingMat_{N}$ of different sizes. The results are shown 
      for predicting with a set of 100 test users based on $P$ random probe 
        ratings each, averaged over 50 runs.  The performance improves with more 
          training users as expected, approaching the performance when training 
          with all users, see~\figref{fig:resultsGroceries}-top for comparison.  
          Given sufficient users in the robot's database ($\geq 750$), we can 
          infer the preferences of new users by assuming fixed object-pair 
          biases and factor vectors without loss in prediction accuracy. Bottom: 
          the prediction error given $P=12$ probe ratings when inferring the 
          preferences of test users given a previously-learned factorization 
          (CF-rand$'$) compared to batch learning with the training and test 
          users combined (CF-rand). As expected, the error for both approaches 
          drops given more training users, converging to 0.20 for 
          $\ratingMat_{750}$ and $\ratingMat_{1000}$, i.e., approaching the 
          performance when training with the full $\ratingMat$, see 
          \figref{fig:resultsGroceries}-bottom.}
  \label{fig:onlineVsOffline}
  \end{figure}

\subsubsection{Learning of New User Preferences}
\label{sec:continualExp}
  In this experiment, we show that our approach is able to learn the preferences 
  of new users based on the object-pair biases and factor vectors learned from 
  previous users, see~\secref{sec:onlineLearning}.  We conducted an experiment 
  similar to that in \secref{sec:GroceriesProbing} using a random probing 
  strategy.  However, we first learned the biases $\biasPair$ and factor vectors 
  $\mathbf{S}$ using rating matrices $\ratingMat_{100}$, $\ratingMat_{250}$, 
  $\dots$, $\ratingMat_{1000}$, corresponding to 100, 250, $\dots$, 1000 
  training users, respectively (\eqref{eq:optimization}). We then used this 
  model to compute the biases $\biasUser$ and factor vectors $\mathbf{T}$ for a 
  set of 100 (different) test users (\eqref{eq:newUserLearning}) and predict 
  their missing ratings. As before, we repeated this experiment for different 
  values $P$ of known probe ratings for the test users.
          The prediction F-score averaged over 50 runs is shown     
          in~\figref{fig:onlineVsOffline}-top. As expected, the performance 
          improves given more training users for learning the $\biasPair$'s and 
          $\mathbf{S}$, converging to the performance when training with all 
          user columns (compare to CF-rand 
              in~\figref{fig:resultsGroceries}-top).
          This validates that, given enough users in the robot's database, we 
          can decouple learning a projection for the object-pairs from the 
          problem of learning the new users' biases and factor vectors.
     
          Moreover, we compared the predictions using this approach (CF-rand$'$) 
          to the standard batch approach that first appends the 
          100 new user columns to the training matrix and learns all biases and 
              factor vectors collaboratively (CF-rand). The prediction error, 
              averaged over the 50 runs, is shown in           
              \figref{fig:onlineVsOffline}-bottom for $P=12$ probe ratings.  As 
              expected, the error for both approaches drops given more training 
              users, converging to 0.20 for $\ratingMat_{750}$ and 
              $\ratingMat_{1000}$, i.e., approaching the performance when 
              training with the full $\ratingMat$ (compare to CF-rand 
                  in~\figref{fig:resultsGroceries}-bottom for $P=12$).  
              Furthermore, with smaller training matrices, we observed a slight 
              advantage in performance for CF-rand$'$. In other words, given 
              fewer ratings, it might be advantageous to solve the smaller 
              optimization problem in~\eqref{eq:newUserLearning}.
     
\paragraph*{Probing and Learning for New Users}
\begin{figure}[]
\setlength{\fboxsep}{0pt}
\centering%
   \hfill\fbox{\includegraphics[width=0.73\columnwidth]{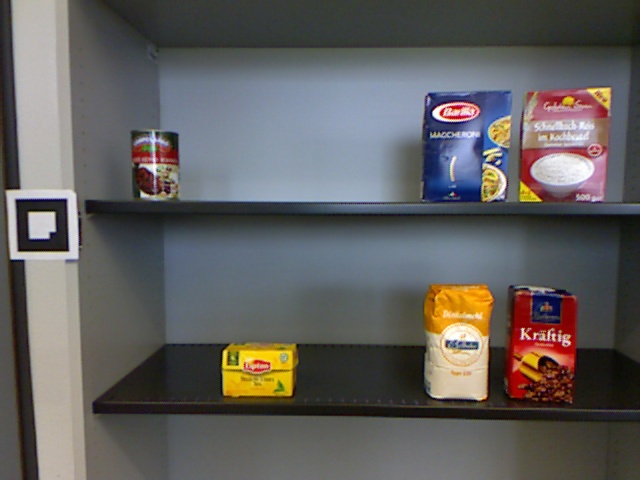}}\hfill~~\\[3mm]
   ~\hfill\fbox{\includegraphics[width=0.73\columnwidth]{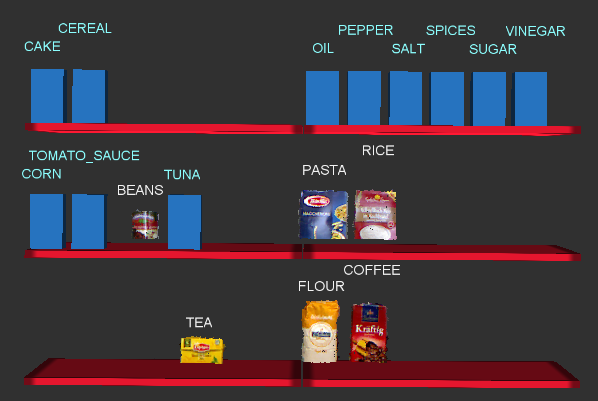}}\hfill~
    \caption{An application of our approach that demonstrates how our 
      predictions for a new user change based on how (probing) objects are 
        arranged on the shelves. Top: the camera image of the scene. To label 
        the objects, we relied on matching SIFT features from a database of 
        images for the used products. Bottom: a visualization of the predicted 
        preferred arrangement of other objects based on the corresponding 
        learned model. Our method is able to infer the user's preferences by 
        adapting to the perceived arrangement. For example, by moving 
        $\mathit{coffee}$ from the shelf containing $\mathit{tea}$ to the one 
        containing $\mathit{flour}$, the predicted arrangement separates       
        $\mathit{cake\ mix}$ and $\mathit{sugar}$ and moves them to different 
        shelves.}\label{fig:shelfPerception}
  \end{figure}
   Using our method, the time for computing the model for one new user (based on 
       a previously-learned factorization) on a consumer-grade notebook was 
   10-20\,ms on average, compared to about 4\,s for batch learning with all 1248 
   user columns ($K=3$).  To demonstrate the applicability of our approach 
   (\secref{sec:cfNewUsers}) in a real-world scenario, we conducted an 
   experiment where we used a Kinect camera to identify a set of objects that we 
   placed on shelves and used the perceived pairwise ratings as probes for 
   inferring a user's preference.  For perception, we clustered the perceived 
   point cloud to segment the objects, and relied on SIFT features matching 
   using a database of product images to label each object. We learned the bias 
   and factors vector for the user associated with this scene using the 
   object-pairs model that we learnt with our crowdsourcing data.  Accordingly, 
   we predicted the pairwise ratings related to objects that are not in the 
   scene and computed the preferred shelves to place them on.  
   \figref{fig:shelfPerception} shows an example where the top image shows the 
   camera image of the scene, and the bottom image shows the corresponding 
   computed shelf arrangement in the rviz visualization environment. Due to 
   physical space constraints, we assume that each shelf is actually divided 
   into two shelves. A video demonstrating how the predicted arrangement changes 
   as the configuration on the shelves varies can be seen at \videourl.

\begin{figure}
\centering
\includegraphics[width=0.8\columnwidth]{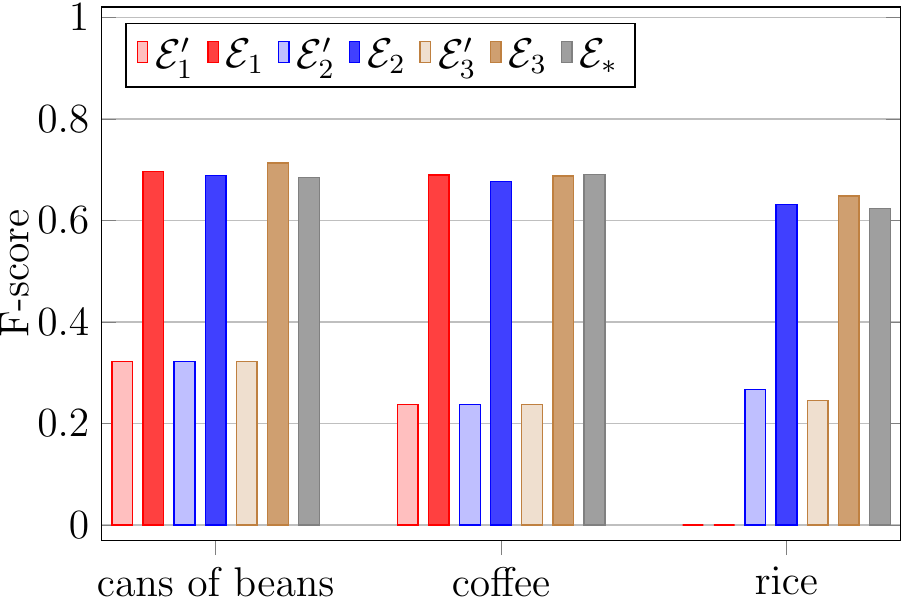}\\[5mm]
\includegraphics[width=0.8\columnwidth]{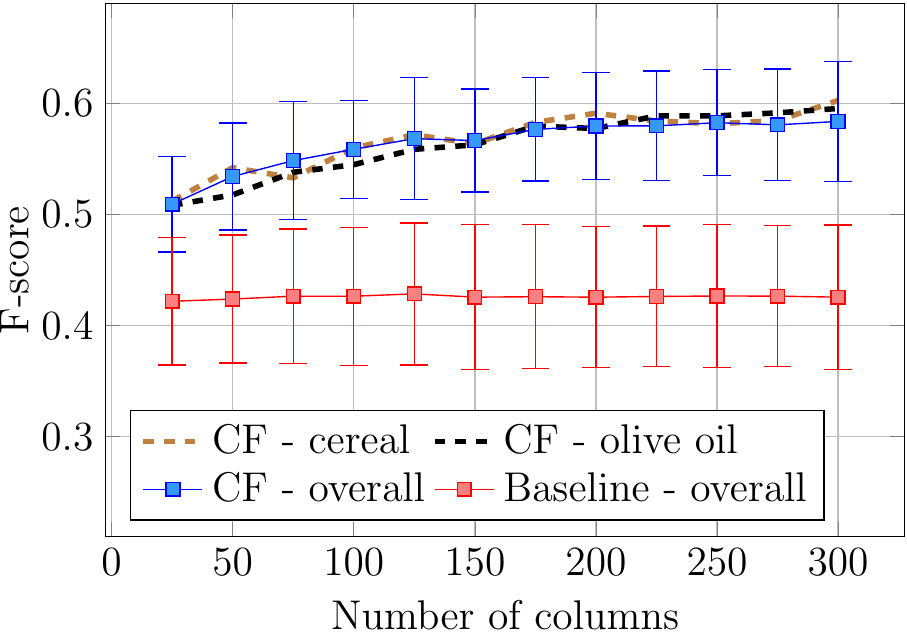}
\caption{Top: we predict preferences related to new objects by using a mixture 
  of experts approach. The experts $\mathcal{E}_1$-$\mathcal{E}_3$ are based on 
    the hierarchies of three online grocery stores. The mixture of experts 
    $\mathcal{E}_*$ is a merged prediction of all three experts based on their
    confidence for a specific user. Therefore, it is able to recover if
    a certain expert cannot find similarities for a new object, as in
    the case of \textit{rice}. The baselines $\mathcal{E}_1'$-$\mathcal{E}_3'$ 
    make predictions based only on the semantic $\mathit{wup}$ similarity of two 
    objects without considering the ratings of similar pairs rated by the user, 
            see~\secref{sec:wup}.  Bottom: the mean F-score for predicting the 
              ratings for a new object vs. the number of training user columns 
              who have rated pairs related to it. As soon as some users have 
              rated pairs related to a new object, our collaborative filtering 
              approach is able to make predictions about it.  The performance 
              improves with more users rating pairs related to the object.}
\label{fig:newGroceries}
\end{figure}

\subsubsection{Predicting Preferences for New Objects}
\label{sec:newGroceries}
In this experiment, we demonstrate that our mixture of experts approach is able 
to make reasonable predictions for previously unrated object-pairs. For this,
we defined three experts by mining the hierarchies of the groceries section of 
three large online stores (amazon.com, walmart.com,
target.com). This includes up to 550 different nodes in the object
hierarchy. For each of the 22 grocery objects, we removed ratings
related to all of its pairs from $\ratingMat$, such that the typical 
collaborative filtering approach cannot make predictions related to that object.  
We used the mixture of experts to predict those ratings using the remaining 
ratings in each column and the expert hierarchies as explained
in~\secref{sec:wup}. The mean F-score over all users for three grocery
objects is shown in~\figref{fig:newGroceries}-top, where the mixture
of experts is denoted by $\mathcal{E}_*$. We also show the individual
expert results ($\mathcal{E}_1$-$\mathcal{E}_3$) and their
corresponding baseline predictions
($\mathcal{E}_1'$-$\mathcal{E}_3'$). The baselines take only the
$\mathit{wup}$ similarity of two objects as the rating of the pair but
do not consider the ratings of similar pairs made by the same user as our 
approach does. As we can see, the results of each individual expert outperform 
the baseline predictions. Note that $\mathcal{E}_*$ is able to overcome the 
shortcomings of the individual experts, as in the case of \textit{rice}. There,
$\mathcal{E}_1$ is unable to find similarities between \textit{rice}
and any of the rated objects, whereas $\mathcal{E}_2$ and
$\mathcal{E}_3$ are able to relate it to \textit{pasta} in their
hierarchies. For two of the objects (\textit{bread} and
\textit{candy}), we were unable to make any predictions, as none of
the experts found similarities between them and other rated
objects. For all other objects, we achieve an average F-score of 0.61.

\begin{figure*}
\centering
\includegraphics[width=0.68\textwidth]{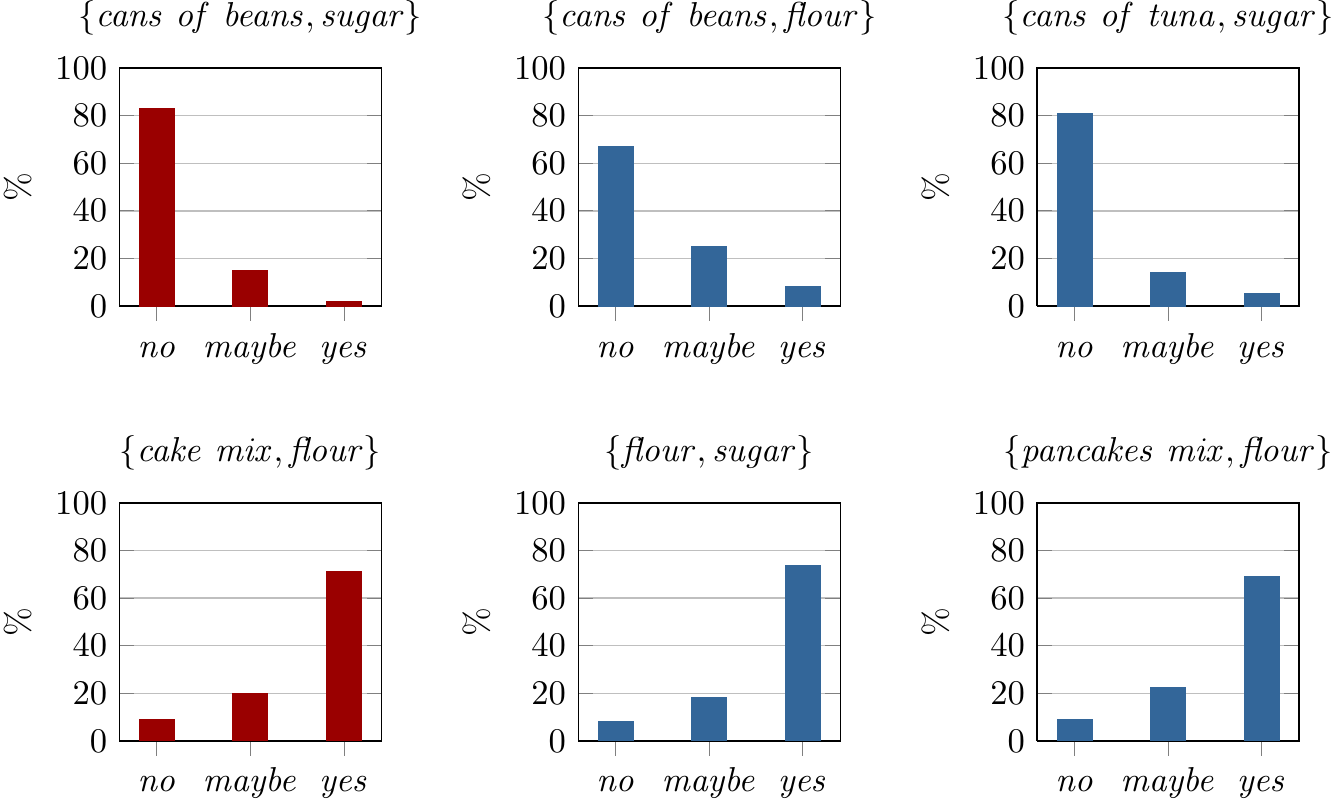}
\caption{The rating distributions depicted in red correspond to example 
  object-pairs that no user had rated in the original crowdsourcing data. We 
    generated those ratings using our mixture of experts approach based on the 
    hierarchies of three online stores. In the case of $\{\mathit{cans\ of\ 
      beans, sugar}\}$, the experts relied on how each user rated similar 
      object-pairs such as $\{\mathit{cans\ of\ beans, flour}\}$ and 
      $\{\mathit{cans\ of\ tuna, sugar}\}$. The rating distributions of those 
      pairs (over all user columns who rated them) are depicted in blue on the 
      same row.  Similarly, in the case of $\{\mathit{cake\ mix, flour}\}$, the 
      experts relied on the known ratings of $\{\mathit{flour, sugar}\}$ and 
      $\{\mathit{pancake\ mix, flour}\}$.}
\label{fig:ratingExamplesGroceriesExperts}
\end{figure*}

\paragraph*{Predicting Ratings for New Object-Pairs}
Furthermore, we applied the same mixture of experts based on the three online 
stores above to extend our ratings matrix $\ratingMat$ with rows for 
object-pairs that no user rated in our crowdsourcing surveys. We created a new 
ratings matrix $\ratingMat'$ of size 214$\times$1284, i.e., with 
35 additional object-pair rows. These included pairs related to the new object 
   $\mathit{cake\ mix}$, as well as other object combinations. For each user 
   column in the original $\ratingMat$, the mixture of experts used 
   already-rated object-pairs to infer ratings for the new pairs.  
   \figref{fig:ratingExamplesGroceriesExperts} shows examples of rating 
   distributions in the resulting ratings matrix for two object-pairs: 
   $\{\mathit{cans\ of\ beans, sugar}\}$ and $\{\mathit{cake\ mix, flour}\}$.  
   Additionally, for each of them, we show the rating distributions of two of 
   their most similar object-pairs that the experts used when making their 
   predictions.  In a later experiment, we use the resulting $\ratingMat'$ with 
   this combination of crowdsourcing and expert-generated ratings to train a 
   factorization model for predicting preferred arrangements of survey 
   participants, see~\secref{sec:shelving}.

\subsubsection{Improvement with Number of Users}
\label{sec:bigData}
We conducted an experiment to show that the performance of our approach improves
with more users in the system. For each object, we removed from $\ratingMat$ all 
columns with ratings related to that object. Over 20 runs, we randomly sampled 
ten different user columns (test users) from these and hid their ratings for 
pairs related to the object. We predicted those ratings using our approach
(\secref{sec:cfLearning}) by incrementally adding more columns of other 
(training) users who rated that object to the ratings matrix in increments
of 25. We evaluated the mean F-score for the predictions for the test
users. The results (CF - overall) are shown in~\figref{fig:newGroceries}-bottom
averaged over 20 different types of objects (those where we had at
least 300 user ratings). We also show the improvement with respect to
two of the objects individually.  The performance of our approach
improves steadily with the number of users who rate pairs related to a
new object, as opposed to a baseline that updates the mean rating over
all users and uses that for predicting. This shows that collaborative
filtering is suitable for lifelong and continual learning of user
preferences.

\subsubsection{Assigning Objects to Shelves Based on Pairwise Preferences}
\label{sec:shelving}
The goal of this experiment is to show that our approach is able to group 
objects into containers to satisfy pairwise preferences, 
        see~\secref{sec:spectralClustering}. We evaluated our approach in two 
        settings.  In the first, we compute the object groupings given ground 
        truth pairwise ratings from users.  In the second, we predict the 
        pairwise ratings according to our approach and use those when grouping 
        the objects on different shelves.

\paragraph{Arrangements Based on Ground Truth Ratings}
We conducted a qualitative evaluation of our approach for grouping objects into 
different containers based on \emph{known} object-pair preferences. We asked a 
group of 16 people to provide their ratings (0 or 1) for 55 object-pairs, 
      corresponding to all pairs for a set of 11 objects. For each participant, 
      we then computed an object arrangement allowing our spectral clustering 
      approach to use up to 6 shelves. We showed each participant the shelf 
      arrangement we computed for them in the rviz visualization environment. We 
      asked each of them to indicate whether they think the arrangement 
      represents their preferences or not.  They then had the choice to make 
      changes to the arrangement by indicating which objects they would move 
      from one shelf to another.  
   
\begin{figure}
   \centering
    \includegraphics[height=4.5cm]{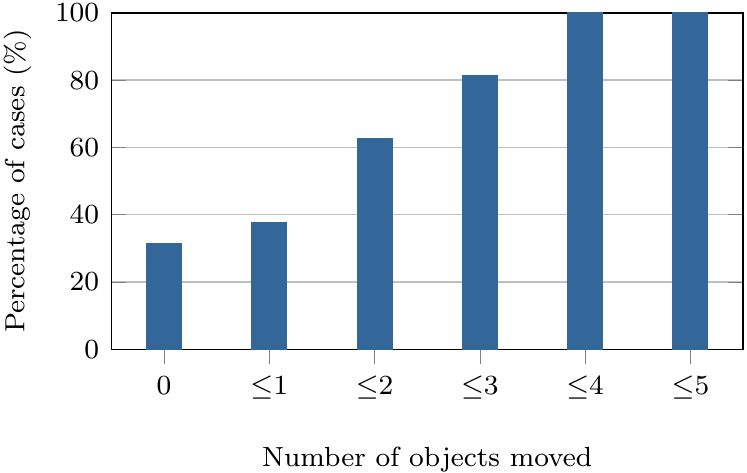}
  \caption{In a small survey with 16 participants, we computed groupings of 
    11 objects based on each participant's ground truth object-pair ratings.  We 
       presented the arrangement to each participant by visualizing it in a 3D 
       visualization environment. Each participant then indicated which objects 
       they would like to move in order to achieve a better arrangement.
      Despite inconsistencies in the pairwise preferences of the participants, 
              we are able to compute reasonable object groupings that correspond 
                to their preferences. On average, the participants modified the 
                locations of only two objects.
  }
  \label{fig:shelvingSurveyDistribution}
\end{figure}

   The results are shown in~\figref{fig:shelvingSurveyDistribution}.
   In five out of 16 cases, the participants accepted the arrangements without 
   any modifications. Overall, the participants modified only two objects on 
   average. Even given ground truth object-pair ratings from the users, there 
   are often several inconsistencies in the preferences that can make it 
   challenging for the eigen-gap heuristic to estimate the best number of 
   shelves to use. Nonetheless, we are able to compute reasonable groupings of 
   objects.  Moreover, the nature of our approach allows the robot to observe 
   arrangements given by users in order to modify its knowledge about their 
   preferences and use this when making new predictions in the future.

\paragraph{Arrangements Based on Predicted Ratings}
\label{sec:shelvingSurvey15}
\begin{figure}
\setlength{\fboxsep}{0pt}
\centering
\fbox{\includegraphics[height=3.5cm]{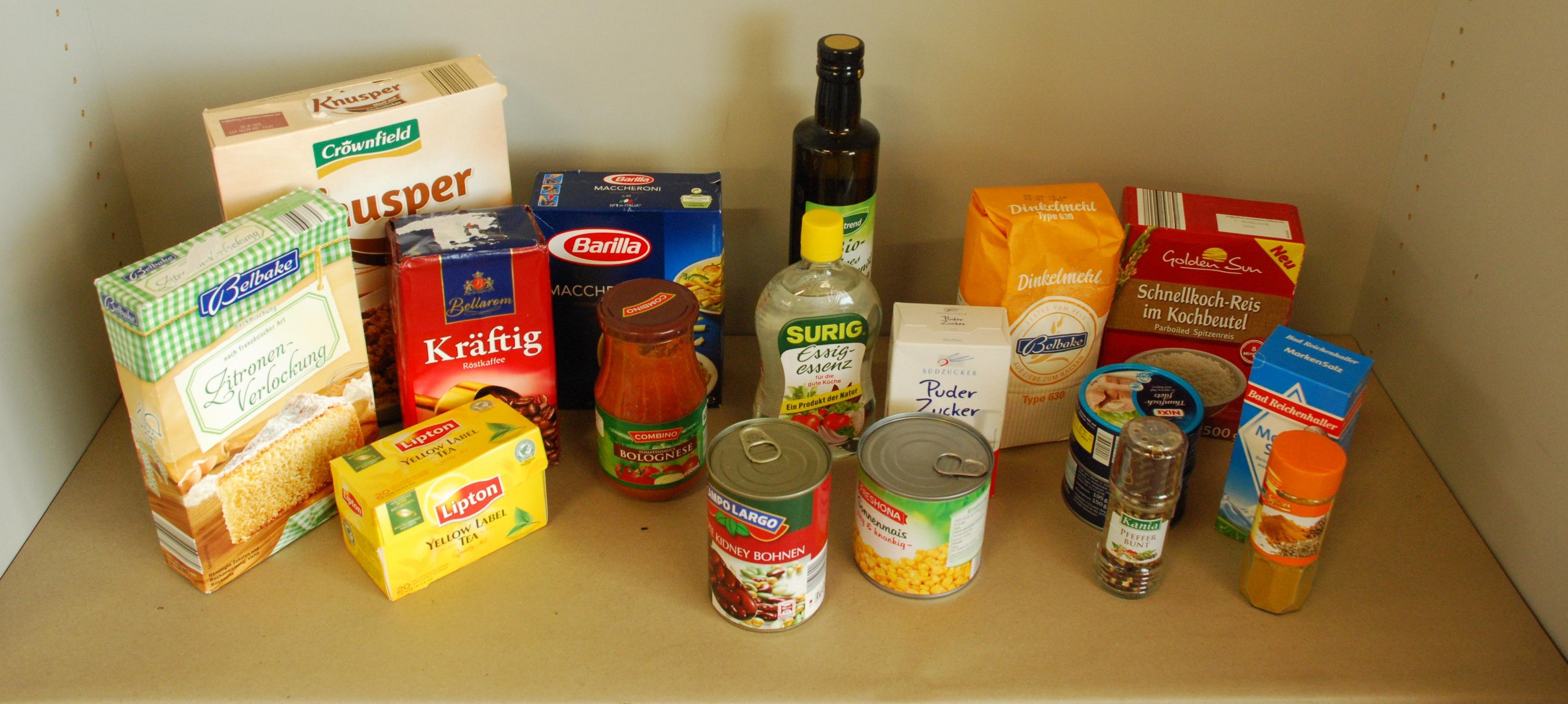}}
\caption{We asked survey participants to organize different types of grocery 
  objects using up to six shelves in order to test our approach for predicting 
    their preferences.}
\label{fig:allGroceries}
\end{figure}

\begin{figure*}[t]
\centering
 \hfill\includegraphics[width=0.3\columnwidth]{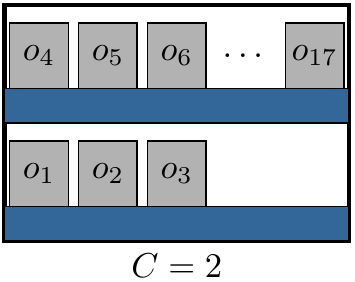}\hfill
 \includegraphics[width=0.3\columnwidth]{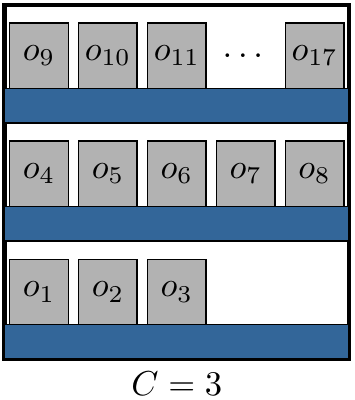}\hfill
 \includegraphics[width=0.3\columnwidth]{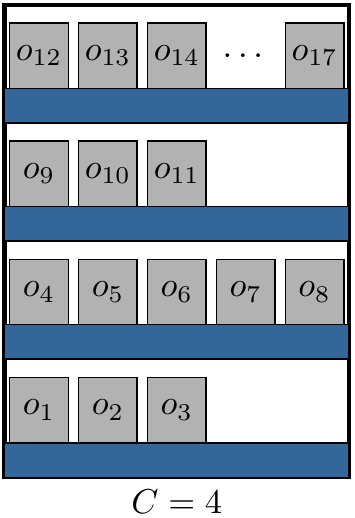}\hfill
 \includegraphics[width=0.3\columnwidth]{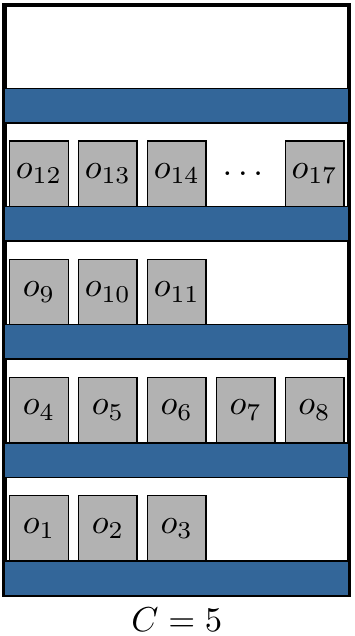}\hfill~
\flushleft
\centering
\begin{tabular}{llllll}
\centering
$o_1:\mathit{cake\ mix}$ & $o_4:\mathit{olive\ oil}$ & $o_7:\mathit{spices}$ & 
$o_{10}:\mathit{coffee}$ & $o_{13}: \mathit{corn}$ & $o_{16}: \mathit{tomato\ 
  sauce}$\\
$o_2:\mathit{flour}$ & $o_5:\mathit{pepper}$ & $o_{8}:\mathit{vinegar}$ & 
$o_{11}:\mathit{tea}$&$o_{14}:\mathit{pasta}$&$o_{17}: \mathit{tuna}$\\
$o_3:\mathit{sugar}$ & $o_6:\mathit{salt}$& $o_{9}:\mathit{cereal}$ & 
$o_{12}:\mathit{beans}$&$o_{15}:\mathit{rice}$&\\
\end{tabular}
\caption{Our approach is able to adapt to the number of containers $C$ available 
  for organizing the objects. In this example, applying the self-tuning 
    heuristic correctly estimates the best number of shelves to use ($C' = 4$), 
              matching the original user preferences. Given more shelves in the 
                scene ($C=5$), our approach still prefers grouping the objects 
                on four shelves, as this maximally satisfies the user's 
                preferences. With only two or three shelves, our method attempts 
                to satisfy the preferences of the user as much as possible by 
                grouping the objects differently.}
\label{fig:shelfAdapting}
\end{figure*}

The goal of this experiment is to evaluate the performance of our approach for 
grouping objects based on \emph{predicted} object-pair ratings.  To collect 
ground truth data for arrangements, we asked 15 people to organize 17 different 
grocery items according to their preferences, using \emph{up to} six shelves, 
        see~\figref{fig:allGroceries}. Four people grouped the items on four 
        shelves, three people used five shelves, and eight people used all six 
        shelves.  \figref{fig:motivation}-left shows examples of arrangements 
        produced by the survey participants. We translated these arrangements to 
        user rating columns with 0 or 1 ratings as described 
        in~\secref{sec:probing}. \figref{fig:shelfAdapting} shows the 
        arrangements our method computes for one of the participants (who used 
            four shelves) when given all ground truth object-pair ratings of 
        that participant. Given four or more shelves to use, we are able to 
        reproduce the original object grouping of this user with $C' = 4$ 
        shelves. The figure also shows how our approach adapts by merging some 
        object groups together when given only two or three shelves to sort the 
        objects.

Our goal is to evaluate the arrangements we compute for the 15 participants 
given only partial knowledge of their ratings, and based on a previously-learned 
model of object-pair biases $\biasPair$ and factor vectors $\mathbf{S}$ from 
training users. To learn the model, we used the ratings matrix $\ratingMat'$ 
described in \secref{sec:newGroceries} above, as this covers all object-pairs 
relevant for the objects in this experiment. For each of the 15 participants, we 
then simulated removing $O$ random objects from their arrangement, and hid all 
ratings related to those objects. Using the remaining ratings as probes, we 
learned the bias $\biasUser$ and factors vector $\mathbf{t}_j$ of each 
participant (\secref{sec:onlineLearning}), and predicted the missing ratings 
accordingly.  Finally, we used those ratings to compute an object arrangement 
for each participant, and compared it to their ground truth arrangement. We 
conducted this experiment for $O = 1, 2, 3, \dots, 12$, repeating it 100 times 
for each setting.

\begin{figure}
\centering
\includegraphics[width=0.7\columnwidth]{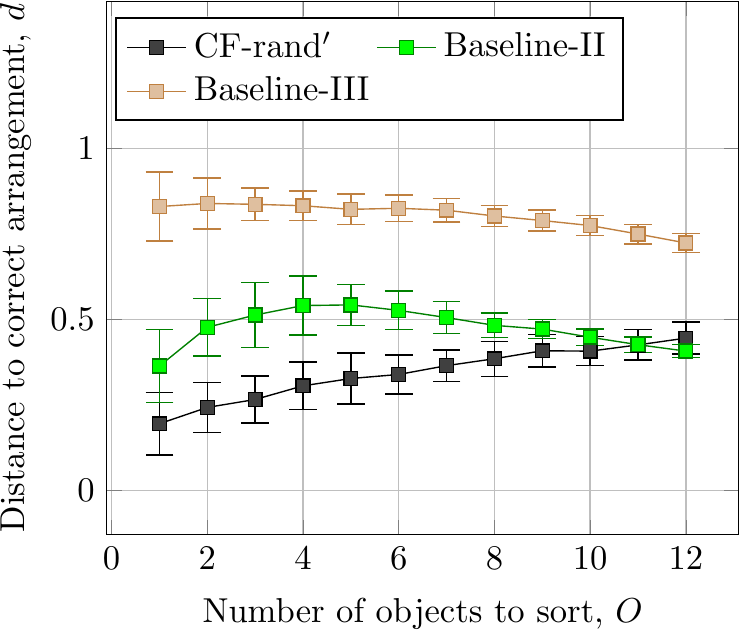}\\[5mm]
\includegraphics[width=0.7\columnwidth]{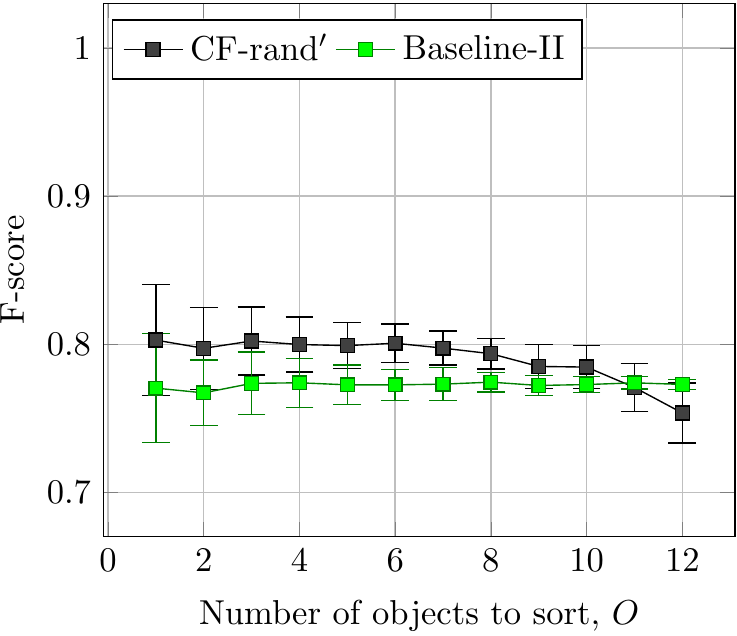}
\caption{We evaluated our approach (CF-rand$'$) for predicting the correct 
  shelves for $O$ randomly-selected objects (out of 17) given the shelves that 
    the remaining objects have been assigned to. CF-rand$'$ predicts the missing 
    object-pair ratings using our approach and then partitions the objects using 
    our spectral clustering method. Baseline-II predicts the missing ratings as 
    the mean rating over all training users, and uses our method    for 
    partitioning the objects. Baseline-III makes no predictions, but randomly 
    assigns each object to one of the six shelves.  Top: the mean error in the 
    computed arrangements (ratio of misplaced objects).  Bottom: the prediction 
    F-score of the object-pair ratings averaged over the 0 and 1 categories. Our 
    approach outperforms both baselines for $O\le10$. As $O$ increases, the 
    error in our predictions increases since there are less probe ratings based 
    on the remaining objects on the shelves to infer the preferences of a user.
}
\label{fig:resultsShelving}
\end{figure}

  We evaluated the results by computing an ``edit" distance~$d$ between the 
  computed and ground truth arrangements. We compute $d$ as the minimum number 
  of objects we need to move from one shelf to another to achieve the correct 
  arrangement, divided by $O$. This gives a normalized error measure between 
  0 and 1 capturing the ratio of misplaced objects. The results (averaged over 
    all runs) are shown in \figref{fig:resultsShelving}-top, where we denote our 
  method by CF-rand$'$. We compared our results to two baselines. The first is 
  Baseline-II, which predicts the missing object-pair preferences using the mean 
  ratings over all users in $\ratingMat'$ and then computes the object groupings 
  using our approach.  The second is Baseline-III, which makes no predictions 
  but simply assigns each object to a random shelf.  
  \figref{fig:resultsShelving}-bottom also shows the mean F-score of our 
  approach and Baseline-II averaged over the 0 and 
  1 rating categories.

  Our approach outperforms both baselines for $O\le10$, and results in a mean 
  error from 0.19 to 0.41, and an F-score from 0.80 to 0.78, as $O$ changes from 
  1 to 10. The model we learned from $\ratingMat'$ (noisy crowdsourcing data 
    from 1284 users augmented by ratings from a mixture of experts) is able to 
  accurately predict the preferences of the new 15 survey participants.  For the 
  same values of $O$, Baseline-II achieves a mean error ranging between   
  0.36 and 0.54, and an F-score of 0.77. As $O$ increases, the error in our 
    predictions increases as expected, since we have less probe ratings based on 
    the remaining objects on the shelves to infer the preferences of a new user.  
    For $O>11$, Baseline-II results in less error than our approach when 
    computing arrangements.  On the other hand, using a random strategy for 
    assigning objects to shelves (Baseline-III) resulted in an error above 0.72 
    for all values of $O$.

\paragraph{Predictions by Humans}
Finally, we conducted a qualitative evaluation to gauge the difficulty of the 
above task for humans. We asked 15 new participants (who did not take part in 
    the previous surveys) to complete partial arrangements by predicting the 
preferences of the 15 users above, whom they do not know. Each participant 
solved six tests.  In each test, we manually reconstructed an arrangement from 
the surveys on the six shelves, then we removed $O$ objects randomly and placed 
them on a nearby table.  We asked each participant to predict the preference of 
another user by inspecting the arrangement in front of them, and to finish 
sorting the remaining $O$ objects accordingly. Each participant solved six such 
cases ($O = \{2, 4, \dots, 12\}$), each corresponding to a different user. As 
before, we computed the error $d$, the ratio of objects that were placed on 
wrong shelves.  The results are shown in~\tabref{tab:HumanPreds}.
When presented with two objects to place only, most participants were able to 
predict the correct shelf for them based on the arrangement of the remaining 
objects. However, given four to twelve objects, the participants misplaced 
between one forth to one third of the objects on average.

\begin{table}
\centering
\normalsize
\caption{The error $d$ in the final arrangement produced by 15~participants when 
  we asked them to sort $O$ objects by predicting the preferences of users they 
    do not know.}
\begin{tabular}{|c|c|}
\cline{1-2} Number of objects, $O$ & Error in arrangement, $d$\\
\cline{1-2}  2  & $0.07\pm0.17$\\
\cline{1-2}  4  & $0.27\pm0.25$\\
\cline{1-2}  6  & $0.24\pm0.22$\\
\cline{1-2}  8  & $0.27\pm 0.18$\\
\cline{1-2}  10  & $0.33\pm0.18$\\
\cline{1-2}  12  & $0.34\pm0.17$\\
\cline{1-2}
\end{tabular}
\label{tab:HumanPreds}
\end{table}

\subsubsection{Real Robot Experiments}
\begin{figure*}
\setlength{\fboxsep}{0pt}
 \centering
~\hfill\fbox{\includegraphics[height=4cm]{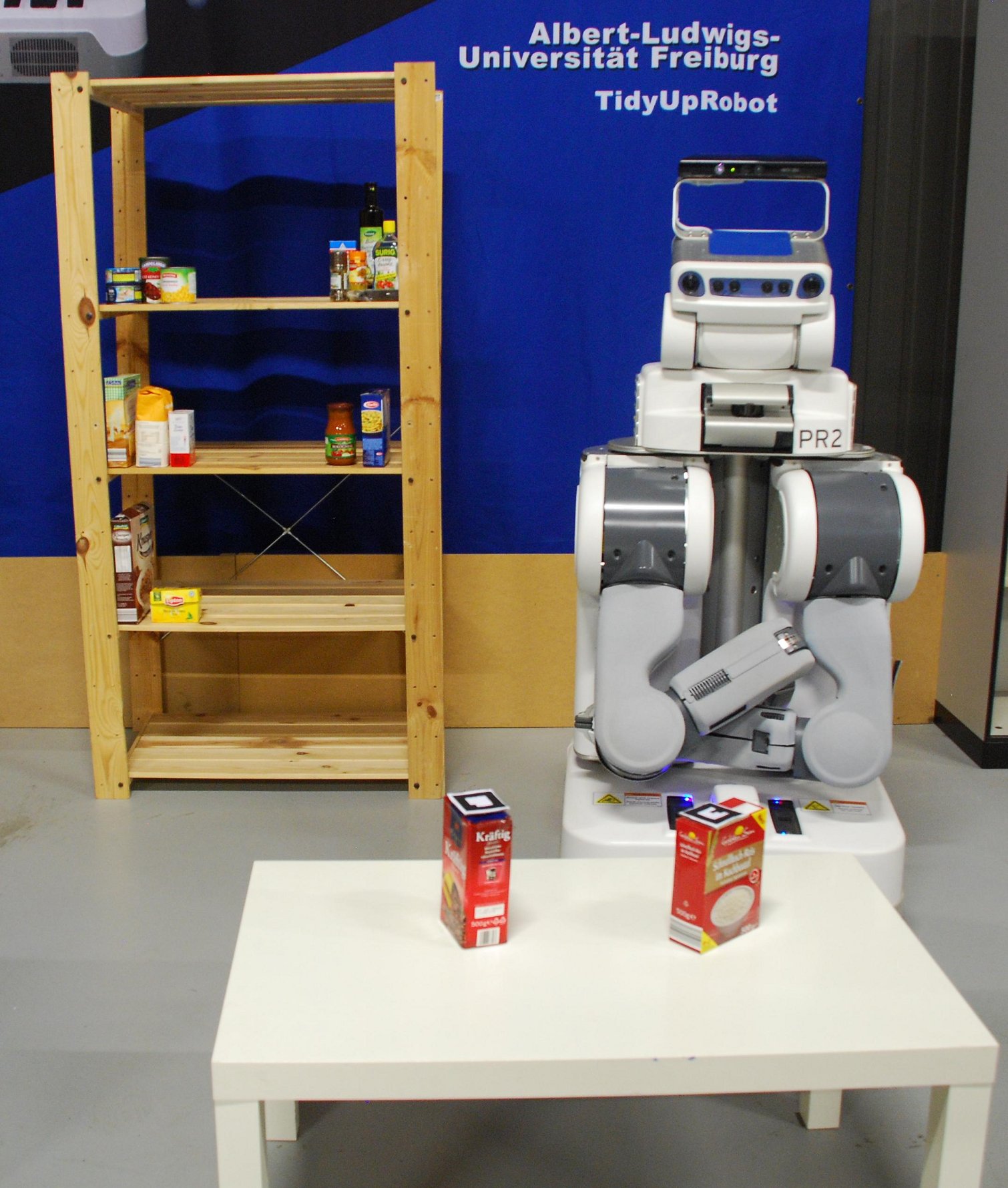}}\hfill
\fbox{\includegraphics[height=4cm]{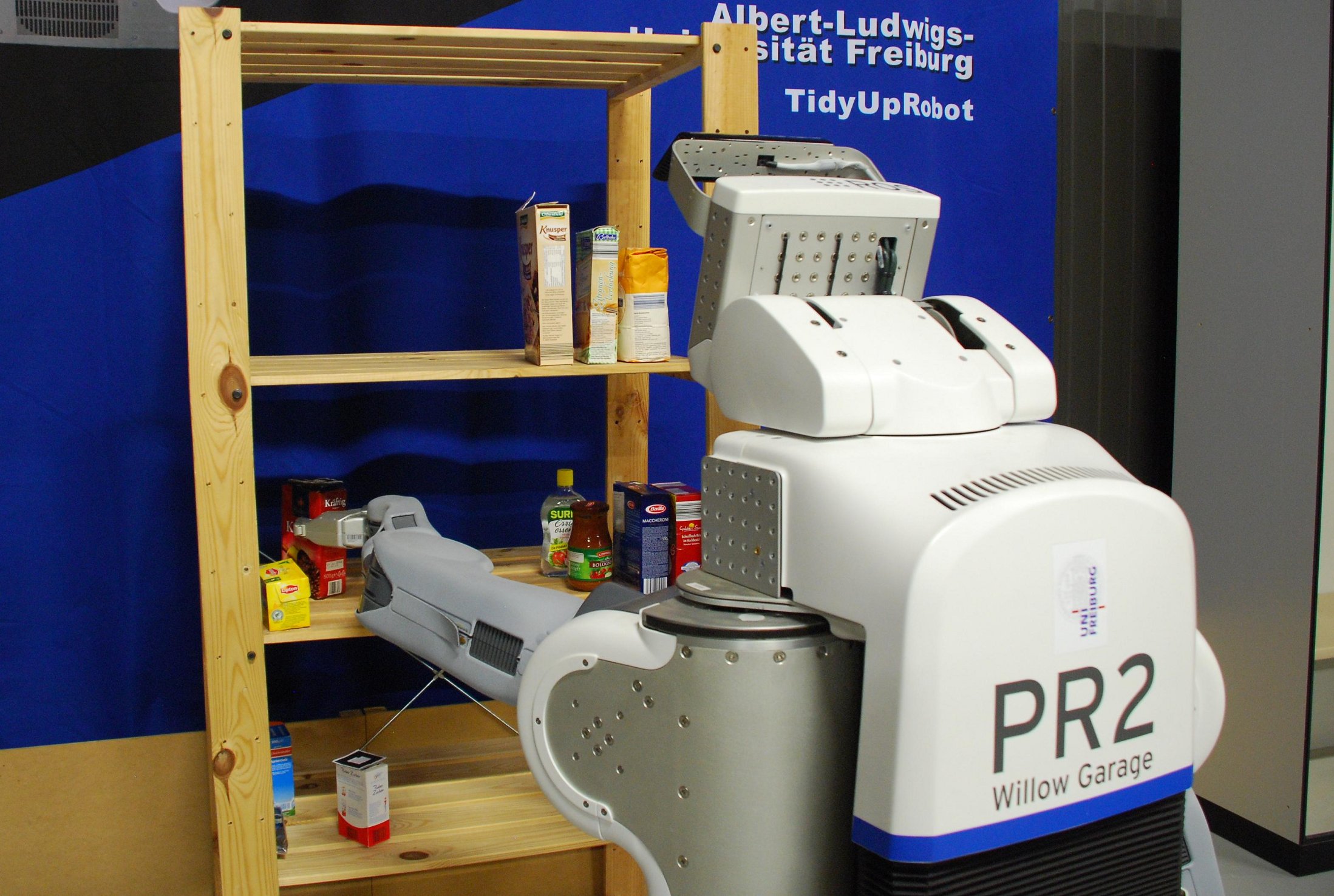}}\hfill~
\caption{Left: the robot has to assign the two objects that are on the table to 
  shelves according to predicted user preferences. In this example, the robot 
    places \textit{coffee} on the same shelf as \textit{tea}, and \textit{rice} 
  next to \textit{pasta}. Right: an example where the robot places 
    \textit{coffee} next to \textit{tea} and \textit{sugar} next to 
    \textit{salt}.}
\label{fig:pr2Exp}
\end{figure*}

We conducted an experiment to illustrate the applicability of our approach 
(\secref{sec:cfLearning}) on a real tidy-up robot scenario using our PR2 robot 
platform, see~\figref{fig:pr2Exp}-left. We executed 25 experimental runs where 
the task of the robot was to fetch two objects from the table and return them to 
their preferred shelves as predicted by our approach. In each run, we arranged 
15 random objects on the shelves according to the preferences of a random user 
   from the survey we conducted in \secref{sec:shelvingSurvey15}, and provided 
   this information as probes for the robot to learn a model of that user 
   (\secref{sec:probing}). The robot used this to predict the pairwise 
   preferences related to the two objects on the table, which it recognized with 
   its Kinect camera using unique fiducial markers we placed on the objects.  It 
   then computed an assignment of the objects to one of the shelves 
   (\secref{sec:spectralClustering}), and expressed those assignments as 
   planning goals in the form of logical predicates in PDDL
     (e.g., $\mathit{on(coffee,\ shelf_2)}$). To achieve those goals, we used a 
     state-of-the-art planner \cite{dornhege13aaaiss} to generate a plan for the 
     robot to navigate to the table, grasp the detected objects, navigate to the 
     shelves, and place the objects on their corresponding shelves (that may be 
         empty or have objects on them) after detecting free space on them. For 
     manipulation, we relied on an out-of-the-box motion planner.  Additionally, 
     we provided the robot with a 3D map of the environment for localizing 
     itself, where we labeled the table and the six shelves.  Overall, the robot 
     predicted the correct shelf assignment for 82$\%$ of the objects using our 
     approach, see~\figref{fig:pr2Exp}-right for an example where the robot 
     successfully placed \textit{coffee} on the same shelf as \textit{tea}.    
     Video excerpts from the experimental runs can be found at \videourl.

\section{Conclusions}
\label{sec:conclusion}
In this work, we presented a novel approach that enables robots to predict user 
preferences with respect to tidying up objects in containers such as shelves or 
boxes. To do so, we first predict pairwise object preferences of the user by 
formulating a collaborative filtering problem. Then, we subdivide the objects in 
containers by modeling and solving a spectral clustering problem. Our approach 
is able to make predictions for new users based on partial knowledge of their 
preferences and a model that we learn collaboratively from several users.  
Furthermore, our technique allows for easily updating knowledge about user 
preferences, does not require complex modeling of objects or users, and improves 
with the amount of user data, allowing for lifelong learning of user 
preferences. To deal with novel objects that the robot encounters, our
approach complements collaborative filtering with a mixture of 
experts based on object hierarchies from the Web.
We trained the system by using surveys from over 1,200 users through 
crowdsourcing, and thoroughly evaluated the effectiveness of our approach for 
two tidy-up scenarios: sorting toys in boxes and arranging groceries on shelves.  
Additionally, we demonstrated the applicability of our approach in a real 
service robot scenario. Our results show that our technique is accurate and is 
able to sort objects into different containers according to user preferences.

\section*{Acknowledgment}
\label{sec:ack}
This work has partly been supported by the German Research Foundation under 
research unit FOR 1513 (HYBRIS) and grant number EXC 1086.

\bibliographystyle{abbrvnat}
\bibliography{refs}

\end{document}